\documentclass[twocolumn,english,compsoc,journal]{IEEEtran}
\usepackage[T1]{fontenc}
\usepackage[latin9]{inputenc}
\usepackage{color}
\usepackage{babel}
\usepackage{verbatim}
\usepackage{amsmath}
\usepackage{amsthm}
\usepackage{amssymb}
\usepackage{graphicx}
\usepackage{subscript}
\makeatletter
\theoremstyle{plain}
\newtheorem{thm}{\protect\theoremname}
\theoremstyle{plain}
\newtheorem{prop}[thm]{\protect\propositionname}

\usepackage[caption=false,font=footnotesize,labelfont=sf,textfont=sf]{subfig}
\usepackage{xcolor} % for text color

\@ifundefined{showcaptionsetup}{}{%
 \PassOptionsToPackage{caption=false}{subfig}}
\usepackage{subfig}
\makeatother

\providecommand{\propositionname}{Proposition}
\providecommand{\theoremname}{Theorem}

\begin{document}

\title{Model-Based Learning for Point Pattern Data}

\author{Ba-Ngu Vo, Dinh Phung, Quang N. Tran, and Ba-Tuong Vo}

\IEEEtitleabstractindextext{
\begin{abstract}
Point patterns are sets or multi-sets of unordered points that arise
in numerous data analysis problems. This article proposes a framework
for model-based point pattern learning using point process theory.
Likelihood functions for point pattern data derived from point process
theory enable principled yet conceptually transparent extensions of
learning tasks, such as classification, novelty detection and clustering,
to point pattern data. Furthermore, tractable point pattern models
as well as solutions for learning and decision making from point pattern
data are developed.
\end{abstract}


}
\maketitle

\section{Introduction\label{subsec:Introduction}}

Point patterns\textendash sets or multi-sets of unordered points\textendash arise
in numerous data analysis problems where they are commonly known as
`bags'. For example, the `bag' in multiple instance learning \cite{amores2013multiple_intance_review,foulds2010multi_instance_review},
the `bag-of-words' in natural language processing and information
retrieval \cite{maron1961NB_setSizeVaried,joachims1996probabilistic,mccallum1998comparison_NBtextClassifi},
the `bag-of-visual-words' in image and scene categorization \cite{csurka2004visual},\cite{fei2005bayesian},
and the `bag-of-features' in sparse data \cite{chickering1999fast_sparse_data,jing2007k_means_sparse_data},
are all point patterns. However, statistical point pattern models
have not received much attention in machine learning for point pattern
data. 

A statistical data model is specified by the \textit{likelihood }function
which can be interpreted as how likely an observation is, given the
parameters of the underlying model. The likelihood can be used to
determine the ``best'' labels for input observations in classification
(supervised learning), the ``best'' cluster parameters in clustering
(unsupervised learning), and outliers in novelty detection (semi-supervised
learning) \cite{markou2003novelty_p1,bishop2006pattern,murphy2012machine}.
The data likelihood function thus plays a fundamental role in model-based
data analysis.

\begin{figure}[h]
\begin{centering}
\vspace{-1mm}
\includegraphics[width=1\columnwidth]{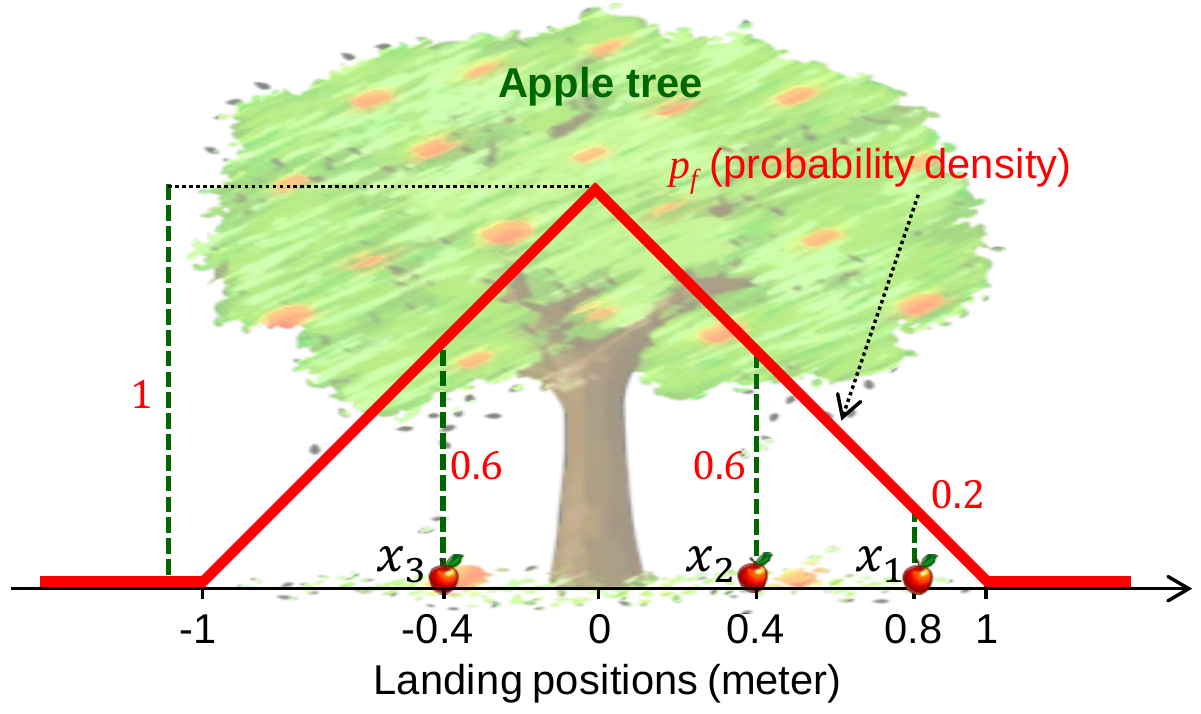}\vspace{-2mm}
\par\end{centering}
\caption{\label{fig:landingPDF}Distribution of landing positions. Position
$x_{1}=0.8\,\mathrm{m}$ is 3 times less likely than $x_{2}=0.4\,\mathrm{m}$
and $x_{3}=-0.4\,\mathrm{m}$ which are equally likely. Credit: clipartbest.com
(apple tree clipart)}
\end{figure}

To motivate the development of suitable likelihood functions for point
patterns, let us consider an example in novelty detection. Suppose
that apples fallen from an apple tree land on the ground independently
from each other, and that the daily point patterns of landing positions
are independent from day to day. Further, the probability density,
$p_{f}$, of the landing position, learned from `normal' training
data, is shown in Fig. \ref{fig:landingPDF}. Since the apple landing
positions are independent, following common practice (see e.g., \cite{maron1961NB_setSizeVaried,joachims1996probabilistic,mccallum1998comparison_NBtextClassifi,csurka2004visual,cadez2000EMclustering_VariableLengthData})
the likelihood that the apples land at positions $x_{1},...,x_{m}$
is given by the joint (probability) density:
\begin{align}
p(x_{1},...,x_{m}) & =\prod_{i=1}^{m}p_{f}(x_{i})\label{eq:NB-likelihood}
\end{align}

Suppose on day 1 we observe one apple landing at $x_{1}$, and on
day 2 we observe two apples landing at $x_{2}$ and $x_{3}$ (see
Fig. \ref{fig:landingPDF}), which of these daily landing patterns
is more likely to be a novelty? Since there is no `novel' training
data in novelty detection, the common practice (see e.g., \cite{markou2003novelty_p1})
is to examine the `normal' likelihoods of the landing patterns
\begin{align*}
p(x_{1}) & =p_{f}(x_{1})=0.2,\\
p(x_{2},x_{3})= & p_{f}(x_{2})\,p_{f}(x_{3})=0.36,
\end{align*}
to identify outliers. Intuitively, the pattern observed on day 1 \emph{is
more likely }to be a novelty since $p(x_{1})<p(x_{2},x_{3})$. However,
had we measured distance in centimeters ($p_{f}$ is scaled by $\mathrm{10}^{-2}$),
then
\[
p(x_{1})=0.002>p(x_{2},x_{3})=0.000036,
\]
thereby, contradicting the previous conclusion! This phenomenon arises
from the incompatibility in the measurement units of the likelihoods
because $p(x_{1})$ is measured in ``$\mathrm{m}^{-1}$'' or ``$\mathrm{cm}^{-1}$''
whereas $p(x_{2},x_{3})$ is measured in ``$\mathrm{m}^{-2}$''
or ``$\mathrm{cm}^{-2}$'', i.e., we are not ``comparing apples
with apples.'' 

\begin{figure}[h]
\begin{centering}
\subfloat[\label{fig:landingPMF_discrete}Distribution of discrete landing positions.]{\begin{centering}
\vspace{-0mm}
\includegraphics[width=1\columnwidth]{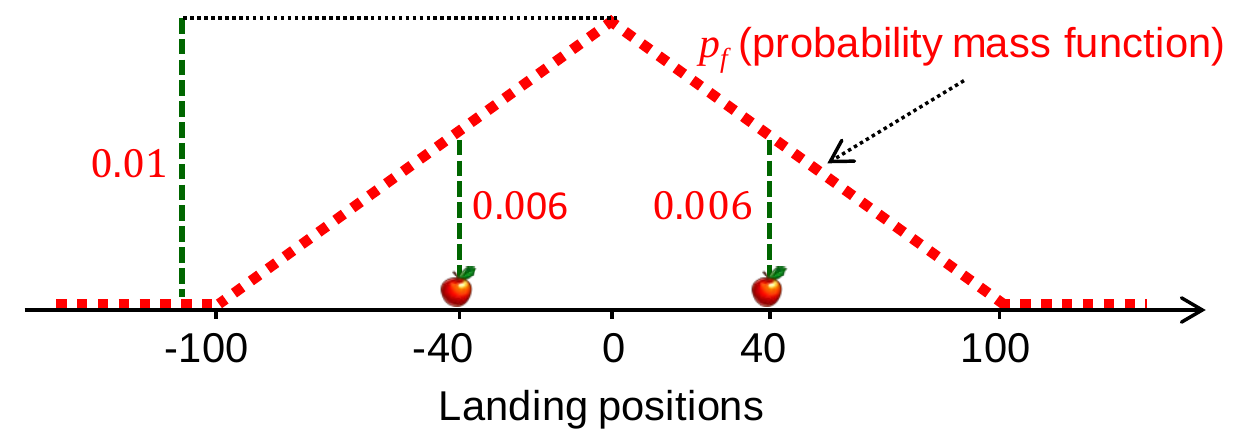}\vspace{-0mm}
\par\end{centering}
\centering{}\vspace{-1mm}
}
\par\end{centering}
\begin{centering}
\subfloat[\label{fig:Normal-data}Examples of 'Normal' observations. ]{\begin{centering}
\includegraphics[bb=0bp -4bp 452bp 85bp,width=1\columnwidth]{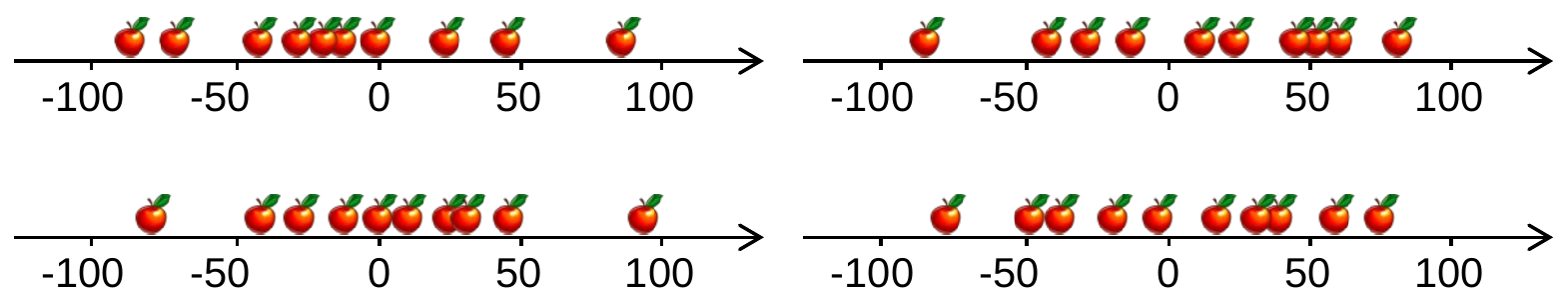}
\par\end{centering}
}\vspace{0mm}
\subfloat[\label{fig:New-data-points}Input observations: $p(X_{1})\approx2\times10^{-23}$
and $p(X_{2})=0.009$.]{\begin{centering}
\includegraphics[bb=0bp -4bp 450bp 52bp,width=1\columnwidth]{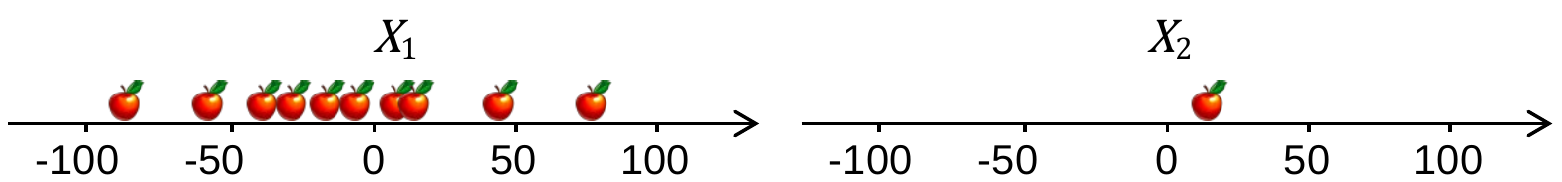}
\par\end{centering}
}
\par\end{centering}
\vspace{-0mm}
\caption{An example with discrete landing positions.}
\end{figure}

The joint density of the landing positions also suffers from another
problem. To eliminate the effect of unit incompatibility, we assume
that there are only 201 positions numbered from $-100$ to $100$,
evenly spaced on the interval $[-1\,\mathrm{m},1\,\mathrm{m}]$. Thus,
instead of a probability density on $[-1\,\mathrm{m},1\,\mathrm{m}]$
we now have a (unit-less) probability mass function on the discrete
set $\left\{ -100,...\,,100\right\} $, as shown in Fig. \ref{fig:landingPMF_discrete}.
Four point patterns from the 'normal' training data set are shown
in Fig. \ref{fig:Normal-data}, while Fig. \ref{fig:New-data-points}
shows 2 new observations $X_{1}$ and $X_{2}$. Since $X_{2}$ has
only 1 feature, whereas $X_{1}$ and the 'normal' observations each
has around 10 features, it is intuitive that $X_{2}$ is novel. However,
its likelihood is much higher than that of $X_{1}$ ($0.009$ versus
$2\times10^{-23}$). This counter intuitive phenomenon arises from\emph{
}the lack of appropriate cardinality information in the likelihood\emph{.} 

The simple example above demonstrates that the joint probability density
of the constituent points is not the likelihood of a point pattern.
In particular, it suffers from incompatibility in the unit of measurement
and does \textit{\emph{not appropriately account for the number of
elements}} in each point pattern. Worse, such inconsistency in a likelihood
function could lead to erroneous results in more complex point pattern
learning tasks. Hence, a proper notion of probability density for
random point pattern is necessary. 

This paper proposes a model-based approach for learning from point
pattern data using point process theory \cite{Stoyan95,Daley88,moller2003point_processes}.
Likelihood functions derived from point process theory are probability
densities of random point patterns, which incorporate both cardinality
and feature information, and avoid the unit of measurement inconsistency.
Moreover, they enable the extension of model-based formulations for
learning tasks such as classification, novelty detection, and clustering
to point pattern data in a conceptually transparent yet principled
manner. Such a framework, facilitates the development of tractable
point pattern models as well as solutions for learning and decision
making. Specifically: 
\begin{itemize}
\item In classification, we propose solutions based on learning point process
models from fully observed training data, and develop an inexpensive
classifier using a tractable class of models;
\item In novelty detection, where observations are ranked according to their
likelihoods, we show that standard point process probability densities
are not suitable for point patterns and develop suitable ranking functions;
\item In clustering we introduce point process mixture models, and develop
an inexpensive Expectation Maximization clustering algorithm for point
pattern using a tractable class of models.
\end{itemize}
These developments have been partially reported in \cite{vo2016model-based_PP,tran2016clustering_PP,PV_RFSMODEL_FUSION14},
respectively. This article provides a more complete study, under a
unified exposition. In Section \ref{sec:Models-for-set-valued} we
review basic concepts from point process theory. Subsequent sections
present the proposed framework for model-based point pattern learning,
in progression from: supervised, namely classification, in Section
\ref{sec:Classification}; semi-supervised, namely novelty detection,
in Section \ref{sec:Anomaly-detection}; to unsupervised, namely clustering,
in Section \ref{sec:Model-Based-MI-Clustering}. Numerical studies
for these learning tasks are presented in Section \ref{sec:Experiments},
followed by some concluding remarks in Section \ref{sec:Conclusions}.

\section{Background \label{sec:Models-for-set-valued}}

Tools for modeling and analysis of point patterns are available from
point process theory, and have proven to be effective in a number
of diverse areas, see e.g., \cite{Stoyan95,illian2008statistical,baccelli_bk10,mahler2014advances}
and references therein. This section outlines the elements of point
process theory and presents some basic models for point pattern data.
For further detail on point processes, we refer the reader to textbooks
such as \cite{Daley88,Stoyan95,moller2003point_processes}. 

\subsection{Point Process\label{subsec:point-process}}

A point pattern is a set or multi-set of unordered points. While a
multi-set is different from a set in that it may contain repeated
elements, a multi-set can also be equivalently expressed as a set.
Specifically, a multi-set with elements $x_{1}$ of multiplicity $N_{1}$,
...., $x_{m}$ of multiplicity $N_{m}$, can be represented as the
set $\{(x_{1},N_{1}),...,(x_{m},N_{m})\}$. A point pattern can be
characterized as a \textit{\textcolor{black}{counting measure}} on
the space $\mathcal{X}$ of features. Given a point pattern $X$,
a counting measure $N$ is defined, for each (compact) set $A\subseteq\mathcal{X}$,
by
\begin{equation}
N(A)=\textrm{number of points of }X\textrm{ falling in }A.
\end{equation}
The values of the counting variables $N(A)$ for all subsets $A$
provide sufficient information to reconstruct the point pattern $X$
\cite{Daley88,Stoyan95}. The points of $X$ are the set of $x$ such
that $N(\{x\})>0$. A point pattern is said to be: \textit{finite}
if it has a finite number of points, i.e., $N(\mathcal{X})<\infty$;
and \textit{simple} if it contains no repeated points, i.e., $N(\{x\})\leq1$
for all $x\in\mathcal{X}$. 

Formally a point process is defined as a \textit{\textcolor{black}{random
counting measure}}. A random counting measure $N$ may be viewed as
a collection of random variables $N(A)$ indexed by $A\subseteq\mathcal{X}$.
A point process is \textit{finite} if its realizations are finite
almost surely, and \textit{\textcolor{black}{simple}} if its realizations
are simple almost surely. 

In this work we are interested in likelihood functions for finite
point patterns. For a countable feature space $\mathcal{X}$, the
likelihood function $f$ is simply the probability of the point pattern.
More concisely,
\begin{equation}
f(\{x_{1},...,x_{i}\})=p_{c}(i)\sum_{\pi}p(x_{\pi(1)},...,x_{\pi(i)}\mid i),
\end{equation}
where: $\pi$ denotes a permutation of $\{1,2,...,i\}$; $p(x_{1},...,x_{i}|i)$
is the joint probability of the features $x_{1},...,x_{i}$, given
that there are $i$ features; and $p_{c}(i)$ is the probability that
there are $i$ features. Conceptually, likelihoods for point patterns
in a countable space is straightforward and requires no further discussion.
Hereon, we consider point processes on a compact subset $\mathcal{X}$
of $\mathbb{R}^{d}$.

\subsection{Probability Density\label{subsec:probability-density}}

The probability density of a point process is the Radon-Nikodym derivative
of its probability distribution with respect to a dominating measure
$\mu$, usually an unnormalised probability distribution of a Poisson
point process.

Let $\nu$ be a (non-atomic $\sigma$-finite) measure on $\mathcal{X}$.
A \textit{Poisson point process} on $\mathcal{X}$ with intensity
measure $\nu$ is a point process such that 
\begin{itemize}
\item for every (compact) set $A\subset\mathcal{X}$, the random variable
$N(A)$ is Poisson distributed with mean $\nu(A)$,
\item if $A_{1},...,A_{m}\subset\mathcal{X}$ are disjoint (compact) sets,
then the random variables $N(A_{1}),...,N(A_{m})$ are independent.
\end{itemize}
In general the probability density of a point process may not exist
\cite{lieshout2000markov,baddeley2007spatial}. To ensure that probability
densities are available, we restrict ourselves to finite point processes
\cite{baddeley2007spatial}. Further, in many applications involving
uncountable feature spaces, the observed point patterns do not have
repeated elements, and hence can be modeled as a simple point process.
A simple finite point process is equivalent to a \textit{random finite
set} \cite{baddeley2007spatial}, i.e., a random variable taking values
in $\mathcal{F}(\mathcal{X})$, the space of finite subsets of $\mathcal{X}$. 

The probability density $f:\mathcal{F}(\mathcal{X})\rightarrow[0,\infty)$
of a random finite set is usually taken with respect to the dominating
measure $\mu$, defined for each (measurable) $\mathcal{T}\subseteq\mathcal{F}(\mathcal{X})$,
by (see e.g., \cite{geyer1999likelihood,moller2003point_processes,vo2005sequential}):
\begin{align}
\mu(\mathcal{T})= & \sum_{i=0}^{\infty}\frac{1}{i!U^{i}}\int\mathbf{1}_{\mathcal{T}}(\{x_{1},...,x_{i}\})d(x_{1},...,x_{i}),\label{eq:commonRefMeasure}
\end{align}
where $U$ is the unit of hyper-volume in $\mathcal{X}$, $\mathbf{1}_{\mathcal{T}}(\cdot)$
is the indicator function for $\mathcal{T}$, and by convention the
integral for $i=0$ is the integrand evaluated at $\emptyset$. The
measure $\mu$ is the unnormalized distribution of a Poisson point
process with unit intensity $1/U$ when $\mathcal{X}$ is bounded.
For this choice of reference measure, it was shown in \cite{vo2005sequential}
that the integral of $f$ is given by 
\begin{align}
\int\negthinspace f(X)\mu(dX)= & \sum_{i=0}^{\infty}\frac{1}{i!U^{i}}\negthinspace\int\negthinspace f(\{x_{1},...,x_{i}\})d(x_{1},...,x_{i}),\label{eq:commonRefMeasure-1}
\end{align}
is equivalent to Mahler\textquoteright s set integral \cite{Mahler_03,mahler2007statistical}
and that densities relative to $\mu$ can be computed using Mahler's
set derivative \cite{Mahler_03,mahler2007statistical}. Note that
the reference measure $\mu$, and the integrand $f$ are all dimensionless. 

The probability density of a random finite set, with respect to $\mu$,
evaluated at $\{x_{1},...,x_{i}\}$ can be written as \cite[p. 27]{lieshout2000markov}
((Eqs. (1.5), (1.6), and (1.7)): 
\begin{equation}
f(\{x_{1},...,x_{i}\})=p_{c}(i)\,i!\,U^{i}f_{i}(x_{1},...,x_{i}),\label{eq:generalRFSdensity}
\end{equation}
where $p_{c}(i)$ is the cardinality distribution, and $f_{i}(x_{1},...,x_{i})$
is a symmetric function\footnote{The notations $f_{m}\left(x_{1},...,x_{m}\right)$ and $f_{m}\left(\{x_{1},...,x_{m}\}\right)$
can be used interchangeably, since $f_{m}$ is symmetric.} denoting the joint probability density of $x_{1},...,x_{i}$ given
cardinality $i$. Note that by convention $f_{0}=1$ and hence $f(\emptyset)=p_{c}(0)$.
It can be seen from (\ref{eq:generalRFSdensity}) that the probability
density $f$ captures the cardinality information as well as the dependence
between the features. Also, $U^{i}$ cancels out the unit of the probability
density $f_{i}(x_{1},...,x_{i})$ making $f$ unit-less, thereby avoids
the unit mismatch. 

\subsection{Intensity and Conditional Intensity}

The \textit{intensity function} $\lambda$ of a point process is a
function on $\mathcal{X}$ such that for any (compact) $A\subset\mathcal{X}$
\begin{equation}
\mathbb{E}\left[N(A)\right]=\int_{A}\lambda(x)dx.\label{eq:generalRFSdensity-2}
\end{equation}
The intensity value $\lambda(x)$ is interpreted as the instantaneous
expected number of points per unit hyper-volume at $x$. 

For a \textit{hereditary} probability density $f$, i.e., $f(X)>0$
implies $f(Y)>0$ for all $Y\subseteq X$, the \textit{conditional
intensity} at a point $u$ is given by \cite{baddeley2007spatial}\vspace{-2mm}
\begin{equation}
\lambda(u,X)=\frac{f(X\cup\{u\})}{f(X)}.\label{eq:generalRFSdensity-1}
\end{equation}
Loosely speaking, $\lambda(u,X)du$ can be interpreted as the conditional
probability that the point process has a point in an infinitesimal
neighbourhood $du$ of $u$ given all points of $X$ outside this
neighbourhood. The intensity function is related to the conditional
intensity by\vspace{-1mm}
\begin{equation}
\lambda(u)=\mathbb{E}\left[\lambda(u,X)\right].\label{eq:generalRFSdensity-2-1}
\end{equation}
For a Poisson point process the conditional intensity equals the intensity 

The probability density of a finite point process is completely determined
by its conditional intensity \cite{Stoyan95,moller2003point_processes}.
Certain point process models are convenient to formulate in terms
of the conditional intensity rather than probability density. Using
the conditional intensity also eliminates the normalizing constant
needed for the probability density. However, the functional form of
the conditional intensity must satisfy certain consistency conditions. 

\subsection{IID cluster model}

Imposing the independence assumption among the features, the model
in (\ref{eq:generalRFSdensity}) reduces to the \emph{IID-cluster}
model \cite{Daley88,Stoyan95}:
\begin{equation}
f(X)=p_{c}(|X|)\,|X|!\,[Up_{f}]^{X},\label{eq:iidRFSdensity}
\end{equation}
where $\left|X\right|$ denotes the cardinality (number of elements)
of $X$, $p_{f}$ is a probability density on $\mathcal{X}$, referred
to as the\emph{ feature density, }and $h^{X}\triangleq\prod_{x\in X}h(x)$,
with $h^{\emptyset}=1$ by convention. Sampling from an IID-cluster
can be accomplished by first sampling the number of points from the
cardinality distribution $p_{c}$, and then sampling the corresponding
number points independently from the feature distribution $p_{f}$
.

When the cardinality distribution $p_{c}$ is Poisson with rate $\rho$
we have the celebrated \textit{\emph{Poisson point process }}\cite{Daley88,Stoyan95}.
\begin{equation}
f(X)=\rho^{|X|}\,e^{-\rho}\,[Up_{f}]^{X}.\label{eq:PoissonPDF}
\end{equation}
The Poisson point process model is completely determined by the intensity
function $\lambda=\rho p_{f}$. Note that the Poisson cardinality
distribution is described by a single non-negative number $\rho$,
hence there is only one degree of freedom in the choice of cardinality
distribution for the Poisson point process model.

\subsection{Finite Gibbs model \label{subsec:Gibbsmodel}}

A well-known general model that accommodates dependence between its
elements is a finite Gibbs process, which has probability density
of the form \cite{Stoyan95,moller2003point_processes}
\begin{equation}
f(X)=\exp\left(V_{0}+\sum_{i=1}^{|X|}\sum_{\{x_{1},...,x_{i}\}\subseteq X}V_{i}(x_{1},...,x_{i})\right),\label{eq:GibbsPDF}
\end{equation}
where $V_{i}$ is called the $i$\textsuperscript{th} potential,
given explicitly by
\[
V_{i}(x_{1},...,x_{i})=\sum_{Y\subseteq\{x_{1},...,x_{i}\}}\negthinspace\negthinspace(-1)^{|\{x_{1},...,x_{i}\}|-|Y|}\log f(Y).
\]
Gibbs models arise in statistical physics, where $\log f(X)$ may
be interpreted as the potential energy of the point pattern. The term
$-V_{1}(x)$ can be interpreted as the energy required to create a
single point at a location $x$, and the term $-V_{2}(x_{1},x_{2})$
can be interpreted as the energy required to overcome the force between
the points $x_{1}$ and $x_{2}$. 

Note that any hereditary probability density of a finite point process
can be expressed in the Gibbs form \cite{baddeley2007spatial}. The
Poisson point process is indeed a first order Gibbs model. Another
example is the hardcore model, where every pair of distinct points
is at least $r$ units apart. In this case, $V_{1}(x)$ is a constant
and
\[
V_{2}(x_{1},x_{2})=\begin{cases}
\begin{array}{cc}
0, & \left\Vert x_{1}-x_{2}\right\Vert >r\\
-\infty, & \left\Vert x_{1}-x_{2}\right\Vert \leq r
\end{array} & .\end{cases}
\]

The next three sections show how point process models are used in
model-based point pattern classification, novelty detection and clustering.

\section{Model-based Classification \label{sec:Classification}}

Classification is the supervised learning task that uses fully-observed
training input-output pairs $\mathcal{D}_{\mathrm{train}}=\{(X_{n},y_{n})\}_{n=1}^{N_{\mathrm{train}}}$
to determine the output class label $y\in\{1,\ldots,K\}$ of each
input observation \cite{bishop2006pattern,murphy2012machine}. This
fundamental machine learning task is the most widely used form of
supervised machine learning, with applications spanning many fields
of study. 

Model-based classifiers for point pattern data have not been investigated.
In multiple instance learning, existing classifiers in the Bag-Space
paradigm are based on distances between point patterns, such as Hausdorff
\cite{huttenlocher1993tracking_Hausdorff,rockafellar2009variational},
Chamfer \cite{gavrila1999Chamfer_distance}, Earth Mover's \cite{zhang2007EMD_kernel,rubner1998EarthMoversDistance}.
Such classifiers do not require any underlying data models and are
simple to use. However, they may perform poorly with high dimensional
inputs due to the curse of dimensionality, and are often computationally
intractable for large datasets \cite{murphy2012machine}, not to mention
that the decision procedure is unclear. On the other hand, knowledge
of the underlying data model can be used to exploit statistical patterns
in the training data, and to devise optimal decision procedures.

Using the notion of probability density for point process from subsection
\ref{subsec:probability-density}, the standard model-based classification
formulation directly extends to point pattern classification:
\begin{itemize}
\item In the \textit{training phase}, we seek likelihoods that ``best''
fit the training data. Specifically, for each $k\in\{1,\ldots,K\}$,
we seek a likelihood function \textbf{$f(\cdot|y=k)$} that best fit
the training input point patterns in $\mathcal{D}_{\mathrm{train}}^{(k)}=\{X:(X,k)\in\mathcal{D}_{\mathrm{train}}\}$,
according to criteria such as\emph{ maximum likelihood} (ML) or Bayes
optimal if suitable priors on the likelihoods are available. 
\item In the \textit{classifying phase}, the likelihoods (learned from training
data) are used to classify input observations. When a point pattern
\textbf{$X$} is passed to query its label, the Bayes classifier returns
the mode of the class label posterior \textbf{$p\left(y=k\mid X\right)$}
computed from the likelihood and the class prior \textbf{$p$} via
Bayes' rule:
\begin{equation}
p\left(y=k\mid X\right)\propto p(y=k)f(X\mid y=k).\label{eq:Bayes-classifier-posterior}
\end{equation}
\end{itemize}
The simplest choices for the class priors are the uniform distribution,
and the categorical distribution, usually estimated from the training
data via \vspace{-1mm}
\[
p(y=k)=\frac{1}{N_{\mathrm{train}}}\sum_{n=1}^{N_{\mathrm{train}}}\delta_{y_{n}}[k],
\]
where $\delta_{i}[j]$ is the Kronecker delta, which takes on the
value 1 when $i=j,$ and zero otherwise. Hence, the main computational
effort in model-based classification lies in the training phase. 

\subsection{Learning Point Process Models}

Learning the likelihood function for class $k$ boils down to finding
the value(s) of the parameter $\theta_{k}$ such that the (parameterized)
probability density $f(\cdot\negthinspace\mid y=k,\theta_{k})$ best
explains the observations $X_{1},...,X_{N}$ in $\mathcal{D}_{\mathrm{train}}^{(k)}$.
In this subsection, we consider a fixed class label and its corresponding
observations $X_{1},...,X_{N}$, and omit the dependence on $k$.

Methods for learning point process models have been available since
the 1970\textquoteright s, see e.g., \cite{moller2003point_processes,baddeley2007spatial}.
We briefly summarize some recognized techniques and presents ML for
IID cluster models as a tractable point pattern classification solution.

\subsubsection{Model fitting via summary statistics}

The method of moments seeks the parameter $\theta$ such that the
expectation of a given statistic of the model point process parameterized
by $\theta$ is equal to the statistic of the observed point patterns
\cite{baddeley2007spatial}. However, this approach is only tractable
when the solution is unique and the expectation is a closed form function
of $\theta$, which is usually not the case in practice, not to mention
that moments are difficult to calculate. 

The method of minimum contrast seeks the parameter $\theta$ that
minimizes some dissimilarity between the expectation of a given summary
statistic (e.g., the K-function) of the model point process and that
of the observed point patterns \cite{baddeley2007spatial}. Provided
that the dissimilarity functional is convex in the parameter $\theta$,
this approach can avoid some of the problems in the method of moments.
However, in general the statistical properties of the solution are
not well understood, not to mention the numerical behaviour of the
algorithm used to determine the minimum.

\subsubsection{Maximum likelihood (ML)}

In the ML approach, we seek the ML estimate (MLE) of $\theta$:
\begin{equation}
\mathrm{MLE}(f(\cdot|\theta);X_{1:N})\triangleq\underset{\theta}{\mbox{argmax}}\left(\prod_{n=1}^{N}f(X_{n}\mid\theta)\right).\label{eq:MLE_general-1}
\end{equation}
The MLE has some desirable statistical properties such as asymptotic
normality and optimality \cite{baddeley2007spatial}. However, in
general, there are problems with non-unique maxima. Moreover, analytic
MLEs are not available because the likelihood (\ref{eq:GibbsPDF})
of many Gibbs models contains an intractable normalizing constant
(which is a function of $\theta$) \cite{moller2003point_processes}. 

To the best of our knowledge, currently there is no general ML technique
for learning generic models such as Gibbs from real data. Numerical
approximation method proposed in \cite{ogata1984likelihood} and Markov
Chain Monte Carlo (MCMC) method proposed in \cite{geyer1994simulation}
are highly specific to the chosen model, computationally intensive,
and require careful tuning to ensure good performance. Nonetheless,
simple models such as the IID-cluster model (\ref{eq:iidRFSdensity})
admits an analytic MLE (see subsection \ref{subsec:ML-Learning-IID}).

Remark: The method of estimating equation replaces the ML estimation
equation
\begin{equation}
\nabla\left(\sum_{n=1}^{N}\log(f(X_{n}\mid\theta))\right)=0\label{eq:MLE_general-1-1}
\end{equation}
by an unbiased sample approximation $\sum_{n=1}^{N}\Psi(\theta,X_{n})=0$
of the general equation $\mathbb{E}_{\theta}\left[\Psi(\theta,X)\right]=0$,
For example, $\Psi(\theta,X_{n})=\nabla\log(f(X_{n}\negthinspace\mid\negthinspace\theta))$
results in ML since it is well-known that (\ref{eq:MLE_general-1-1})
is an unbiased estimating equation. Setting $\Psi(\theta,X_{n})$
to the difference between the empirical value and the expectation
of the summary statistic results in the method of moments. Takacs-Fiksel
is another well-known family of estimating equations \cite{takacs1986estimator,fiksel1988estimation}.

\subsubsection{Maximum Pseudo-likelihood}

Maximum pseudo-likelihood (MPL) estimation is a powerful approach
that avoids the intractable normalizing constant present in the likelihood
while retaining desirable properties such as consistency and asymptotic
normality in a large-sample limit \cite{besag1975statistical,besag1977some}.
The key idea is to replace the likelihood of a point process (with
parameterized conditional intensity $\lambda_{\theta}(u;X)$) by the
pseudo-likelihood:
\begin{equation}
\mathrm{PL}(\theta;X_{1:N})=\prod_{n=1}^{N}e^{-\int\lambda_{\theta}(u;X_{n})du}\left[\lambda_{\theta}(\cdot;X_{n})\right]^{X_{n}}.\label{eq:condintensity-1-1}
\end{equation}
The rationale behind this strategy is discussed in \cite{besag1975statistical}.
Up to a constant factor, the pseudo-likelihood is indeed the likelihood
if the model is Poisson, and approximately equal to the likelihood
if the model is close to Poisson. The pseudo-likelihood may be regarded
as an approximation to the likelihood which neglects the inter-point
dependence.

An MPL algorithm has been developed by Baddeley and Turner in \cite{baddeley2000practical}
for point processes with sufficient generality such as Gibbs whose
conditional intensity has the form 
\[
\lambda(u,X)=\exp\negthinspace\left(\sum_{i=1}^{|X|+1}\negthinspace\sum_{\{x_{1},...,x_{i-1\negthinspace}\}\subseteq X}\negthinspace\negthinspace\negthinspace V_{i}(u,x_{1},...,x_{i-1})\right).
\]
By turning the pseudo-likelihood of a general point process into a
classical Poisson point process likelihood, MPL can be implemented
with standard generalized linear regression software \cite{baddeley2000practical}.
Due to its versatility, the Baddeley-Turner algorithm is the preferred
model fitting tool for point processes. 

The main hurdle in the application of the Baddeley-Turner algorithm
to point pattern classification is the computational requirement.
While this may not be an issue in spatial statistics applications,
the computational cost is still prohibitive with large data sets often
encountered in machine learning. On the other hand, disadvantages
of MPL (relative to ML) such as small-sample bias and inefficiency
\cite{besag1977some,jensen1991pseudolikelihood} become less significant
with large data. Efficient algorithms for learning general point process
models is an on going area of research.

\subsubsection{ML Learning for IID-Clusters\label{subsec:ML-Learning-IID}}

Computationally efficient algorithms for learning point process models
are important because machine learning usually involve large data
sets (compared to applications in spatial statistics). Since learning
a general point process is computationally prohibitive, the IID-cluster
model (\ref{eq:iidRFSdensity}) provides a good trade-off between
tractability and versatility by neglecting interactions between the
points.

Since an IID-cluster model is uniquely determined by its cardinality
and feature distributions, we consider a parameterization of the form:
\begin{align}
f(X\mid\xi,\varphi)= & p_{\xi}(|X|)\,|X|!\,U^{|X|}p_{\varphi}^{X},\label{eq:iidRFSdensity_wParams}
\end{align}
where $p_{\xi}$ and $p_{\varphi}$, are the cardinality and feature
distributions parameterized by $\xi$ and $\varphi$, respectively.
Learning the underlying parameters of an IID-cluster model amounts
to estimating the parameter $\theta=(\xi,\varphi)$ from training
data. 

The form of the IID-cluster likelihood function allows the MLE to
separate into the MLE of the cardinality parameter $\xi$ and MLE
of the feature parameter $\varphi$. This is stated more concisely
in Proposition \ref{prop:IID-cluster} (the proof is straightforward,
but included for completeness).
\begin{prop}
\label{prop:IID-cluster}Let $X_{1},...,X_{N}$ be $N$ i.i.d. realizations
of an IID-cluster with parameterized cardinality distribution $p_{\xi}$
and feature density $p_{\varphi}$. Then the MLE of $(\xi,\varphi),$
is given by
\begin{eqnarray}
\hat{\xi} & = & \mathrm{MLE}\left(p_{\xi};|X_{1}|,...,|X_{N}|\right),\label{eq:Prop1_1}\\
\hat{\varphi} & = & \mathrm{MLE}\left(p_{\varphi};\uplus_{n=1}^{N}X_{n}\right),\label{eq:Prop1_2}
\end{eqnarray}
where $\uplus$ denotes disjoint union.
\end{prop}
\begin{IEEEproof}
Using (\ref{eq:iidRFSdensity_wParams}), we have
\begin{align*}
\prod_{n=1}^{N}f(X_{n}\mid\xi,\varphi) & =\prod_{n=1}^{N}p_{\xi}(|X_{n}|)\,|X_{n}|!\,U^{|X_{n}|}p_{\varphi}^{X_{n}}\\
 & =\prod_{n=1}^{N}|X_{n}|!\,U^{|X_{n}|}\prod_{n=1}^{N}p_{\xi}(|X_{n}|)\prod_{n=1}^{N}p_{\varphi}^{X_{n}}
\end{align*}
Hence, to maximize the likelihood we simply maximize the second and
last products in the above separately. This is achieved with (\ref{eq:Prop1_1})
and (\ref{eq:Prop1_2}).
\end{IEEEproof}
Observe from Proposition \ref{prop:IID-cluster} that the MLE of the
feature density parameter is identical to that used in NB. For example:
if the feature density is a Gaussian, then the MLEs of the mean and
covariance are
\begin{align}
\hat{\mu} & =\frac{1}{N}{\textstyle \sum}_{n=1}^{N}{\textstyle \sum}_{x\in X_{n}}x,\\
\hat{\Sigma} & =\frac{1}{N}{\textstyle \sum}_{n=1}^{N}{\textstyle \sum}_{x\in X_{n}}\left(x-\hat{\mu}\right)\left(x-\hat{\mu}\right)^{\mathrm{T}};
\end{align}
if the feature density is a Gaussian mixture, then the MLE of the
Gaussian mixture parameters can be determined by the EM algorithm.
Consequently, learning the IID-cluster model requires only one additional,
but relatively inexpensive, task of computing the MLE of the cardinality
parameters. 

For a categorical cardinality distribution, i.e., $\xi=\left(\xi_{1},...,\xi_{M}\right)$
where $\xi_{k}=\mbox{Pr}(|X|=k)$ and $\sum_{k=1}^{M}\xi_{k}=1$,
the MLE of the cardinality parameter is given by
\begin{eqnarray}
\hat{\xi}_{k} & = & \frac{1}{N}{\textstyle \sum}_{n=1}^{N}\delta_{k}[|X_{n}|].
\end{eqnarray}
Note that to avoid over-fitting, the standard practice of placing
a Laplace prior on the cardinality distribution can be applied, i.e.
replacing the above equation by $\hat{\xi}_{k}\propto\epsilon+{\textstyle \sum}_{n=1}^{N}\delta_{k}[|X_{n}|]$,
where $\epsilon$ is a small number. 

For a Poisson cardinality distribution parameterized by the rate $\xi=\rho$,
the MLE is given by
\begin{equation}
\hat{\rho}=\frac{1}{N}{\textstyle \sum}_{n=1}^{N}|X_{n}|.
\end{equation}
It is also possible to derive MLEs for other families of cardinality
distributions such as Panjer, multi-Bernoulli, etc. 

Remark: Proposition \ref{prop:IID-cluster} also extends to Bayesian
learning for IID-clusters if the prior on $(\xi,\varphi)$ separates
into priors on $\xi$ and $\varphi$. Following the arguments in the
proof of Proposition \ref{prop:IID-cluster}, the maximum aposteriori
(MAP) estimate of $(\xi,\varphi)$ separates into MAP estimates of
$\xi$ and $\varphi$. Typically a (symmetric) Dirichlet distribution
$Dir(\cdot|\eta/K,...,\eta/K)$, with dispersion $\eta$ on the unit
$M$-simplex, can be used as a prior on the categorical cardinality
distribution. The prior for $\varphi$ depends on the form of the
feature density $p_{\varphi}$ (see also subsection \ref{subsec:infinite-mixture}
for conjugate priors of the Poisson model). Indeed, Bayesian learning
for point process models can be also be considered as a variation
of the Bayesian point pattern clustering problem in Section \ref{sec:Model-Based-MI-Clustering}.

\section{Model-based Novelty Detection\label{sec:Anomaly-detection}}

Novelty detection is the semi-supervised task of identifying observations
that are significantly different from the rest of the data \cite{markou2003novelty_p1,pimentel2014review_novel_detect}.
In novelty detection, there is no novel training data, only `normal'
training data is available. Hence it is not a special case of classification
nor clustering \cite{chandola2009anomaly,hodge2004survey}, and is
a separate problem in its own right. 

Similar to classification, novelty detection involves a training phase
and a detection phase. Since novel training data is not available,
input observations are ranked according to how well they fit the `normal'
training data and those not well-fitted are deemed novel or anomalous
\cite{chandola2009anomaly,hodge2004survey}. The preferred measure
of goodness of fit is the `normal' likelihoods of the input data.
To the best of our knowledge, there are no novelty detection solutions
for point pattern data in the literature. 

In this section we present a model-based solution to point pattern
novelty detection. The training phase in novelty detection is the
same as that for classification. However, in the detection phase the
ranking of likelihoods is not applicable to point pattern data, even
though point process probability density functions are unit-less and
incorporates both feature and cardinality information. In subsection
\ref{sec:PDF-likelihood}, we discuss why such probability densities
are not suitable for ranking input point patterns, while in subsection
\ref{sec:ranking-function} we propose a suitable ranking function
for novelty detection.

\subsection{Probability density and likelihood\label{sec:PDF-likelihood}}

This subsection presents an example to illustrate that the probability
density of a point pattern does not necessarily indicate how likely
it is. For this example, we reserve the term \textit{likelihood} for
the measure of how likely or probable a realization is. 

\begin{figure}[h]
\begin{centering}
\vspace{-4mm}
\par\end{centering}
\begin{centering}
\subfloat[\label{fig:feat-dist-fat} `Short' uniform density]{\begin{centering}
\hspace{-3mm}\includegraphics[width=0.48\columnwidth]{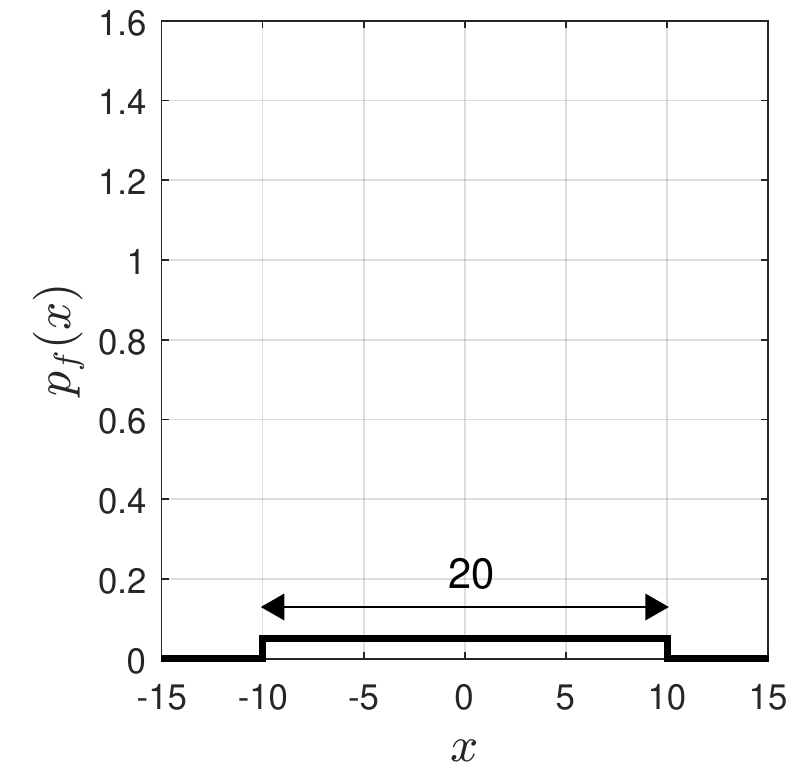}
\par\end{centering}
}~~~~\subfloat[\label{fig:feat-dist-thin}`Tall' uniform density ]{\centering{}\hspace{-3mm}\includegraphics[width=0.47\columnwidth]{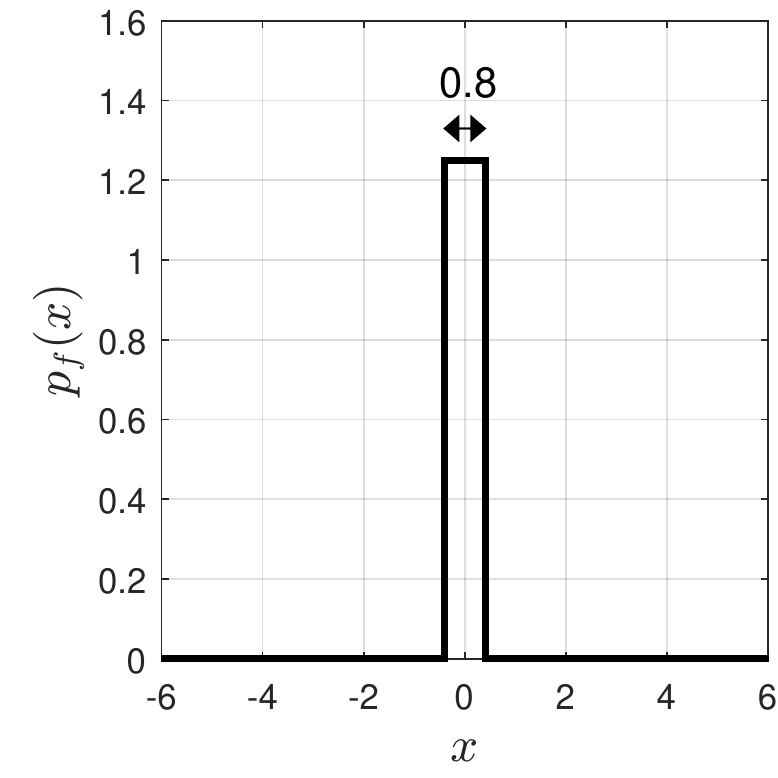}}
\par\end{centering}
\begin{centering}
\subfloat[\label{fig:Card-dist}Cardinality distribution on \{0,...,20\} with
one mode at 10, the remaining cardinalities are equally likely with
total mass 0.2. ]{\begin{centering}
\hspace{18mm}\includegraphics[width=0.48\columnwidth]{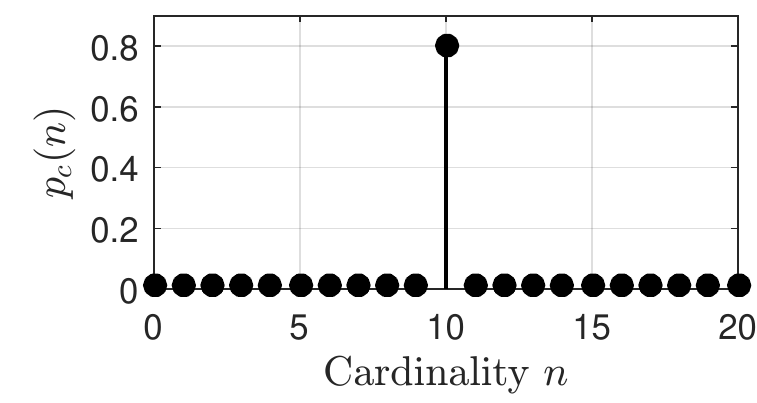}\hspace{18mm}
\par\end{centering}
}
\par\end{centering}
\caption{\label{fig:card_feat_dis}Feature and cardinality distributions for
2 IID-clusters. }
\vspace{-1mm}
\end{figure}

Consider two IID-cluster models with different uniform feature densities
and a common cardinality distribution as shown in Fig. \ref{fig:card_feat_dis}.
Due to the uniformity of $p_{f}$, it follows from (\ref{eq:iidRFSdensity})
that point patterns from each IID-cluster model with the same cardinality
have the same probability density. Note from \cite{Daley88} that
to sample from an IID-cluster model, we first sample the number of
points from the cardinality distribution, and then sample the corresponding
number of points independently from the feature distribution. For
an IID-cluster model with uniform feature density, the joint distribution
of the features is completely uninformative (total uncertainty) and
so the likelihood of a point pattern should be proportional to the
probability of its cardinality.

\begin{figure}[h]
\begin{centering}
\vspace{-3mm}
\subfloat[\label{fig:RFSdens_prob-fat}`Short' uniform feature density]{\begin{centering}
\negmedspace{}\negmedspace{}\negmedspace{}\negmedspace{}\negmedspace{}\includegraphics[width=0.48\columnwidth]{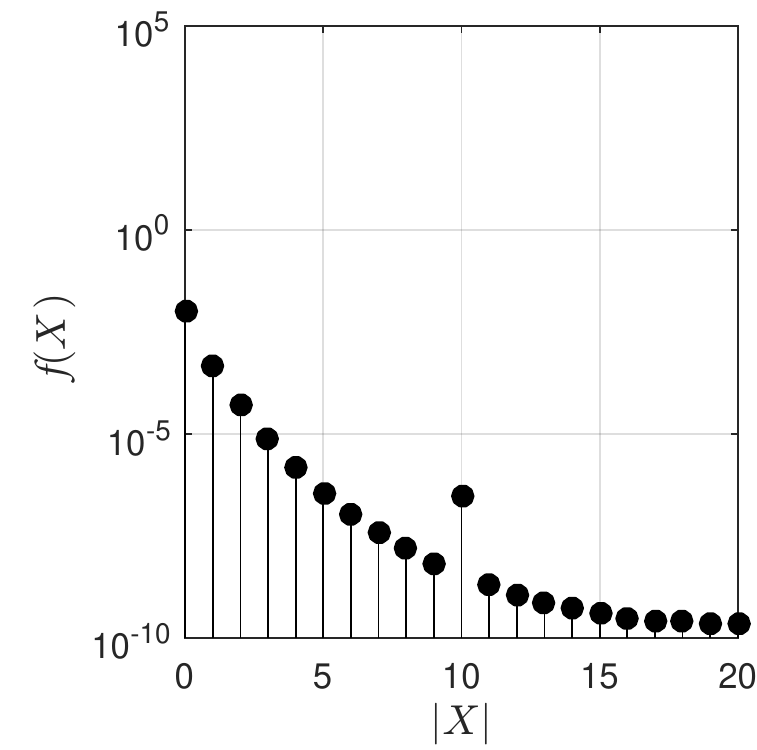}
\par\end{centering}
}~\subfloat[\label{fig:RFSdens_prob-thin}`Tall' uniform feature density]{\centering{}\includegraphics[width=0.48\columnwidth]{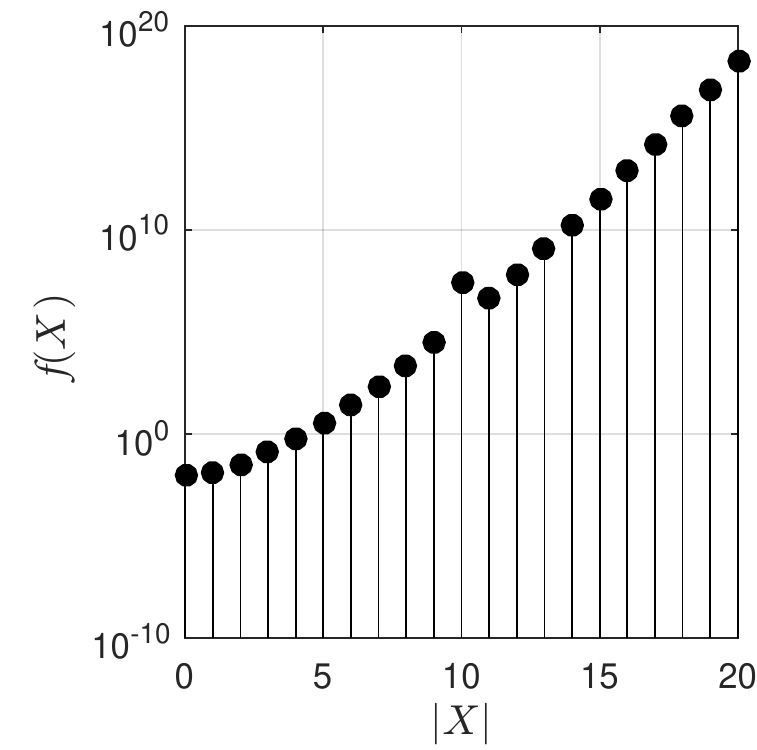}}
\par\end{centering}
\caption{\label{fig:RFSdens_prob}Probability density vs cardinality for 2
IID-clusters. }
\vspace{-1mm}
\end{figure}

If the probability density were an indication of how likely a point
pattern is, then the plot of probability density against cardinality
should resemble the cardinality distribution. However, this is not
the case. Fig. \ref{fig:RFSdens_prob} indicates that for the IID-cluster
with `short' feature density, the probability density tends to decrease
with increasing cardinality (Fig. \ref{fig:RFSdens_prob-fat}). This
phenomenon arises because the feature density given cardinality $n$
is $(1/20)^{n}$, which vanishes faster than the $n!$ growth (for
$n\leq20)$. The converse is true for the IID-cluster with `tall'
feature density (Fig. \ref{fig:RFSdens_prob-thin}). Thus, point patterns
with highest/least probability density are not necessarily the most/least
probable. 

\begin{figure}[h]
\begin{centering}
\vspace{-4mm}
\subfloat[\label{fig:1-D-Gaussian}]{\begin{centering}
\includegraphics[bb=4bp 0bp 188bp 187bp,width=0.46\columnwidth]{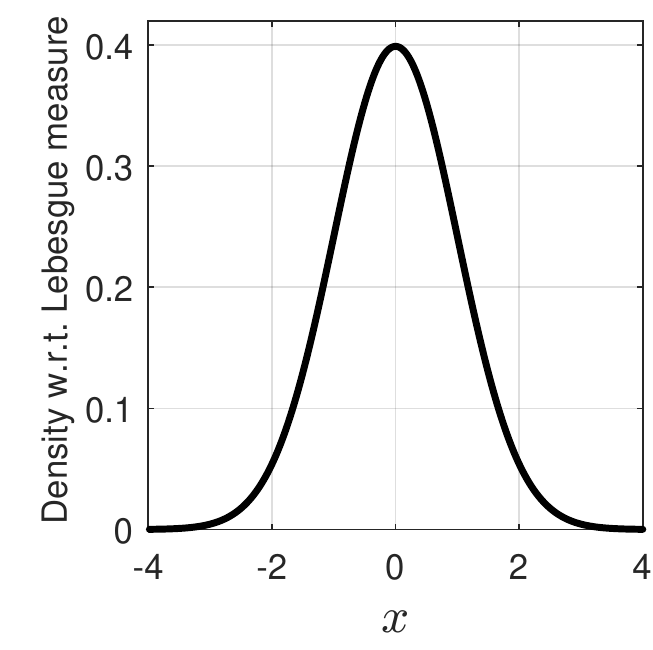}
\par\end{centering}

}~\subfloat[\label{fig:Gau_wrt_Gau}]{\begin{centering}
\includegraphics[bb=4bp 0bp 188bp 187bp,width=0.46\columnwidth]{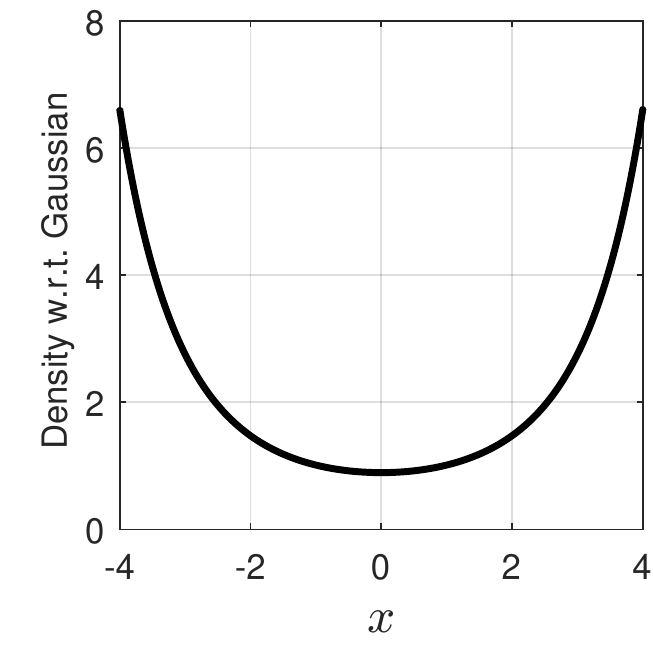}
\par\end{centering}

}
\par\end{centering}
\caption{Density of a zero-mean unit-variance Gaussian w.r.t.: (a) Lebesgue
measure; (b) zero-mean Gaussian with variance $0.8$. }
\vspace{-1mm}
\end{figure}

Such problem arises from the non-uniformity of the reference measure.
A measure $\mu$ is said to be uniform if for any measurable region
$A$ with $\mu(A)<\infty$, all points of $A$ (except on set of measure
zero) are equi-probable under the probability distribution $\mu/\mu(A)$.
One example is the Lebesgue measure $vol$ on $\mathbb{R}^{n}$: given
any bounded measurable region $A$, all realizations in $A$ are equally
likely under the probability distribution $vol(\cdot)/vol(A)$. The
probability density $f(X)=P(dX)/\mu(dX)$ (as a Radon-Nikodym derivative)
at a point $X$ is the ratio of probability measure to reference measure
at an infinitesimal neighbourhood of $X$. Hence, unless the reference
measure is uniform, $f(X)$ is not a measure of how likely $X$ is.
This is also true even for probability densities on the real line.
For example, the probability density of a zero-mean Gaussian distribution
with unit variance relative to the (uniform) Lebesgue measure is the
usual Gaussian curve shown in Fig. \ref{fig:1-D-Gaussian}, while
its density relative to a zero-mean Gaussian distribution with variance
0.8 is shown in Fig. \ref{fig:Gau_wrt_Gau}, where the most probable
point has the least probability density value. 

The reference measure $\mu$ defined by (\ref{eq:commonRefMeasure})
is not uniform because for a bounded region $\mathcal{T}\subseteq\mathcal{F}(\mathcal{X})$,
the probability distribution $\mu/\mu(\mathcal{T})$ is not necessarily
uniform (unless all points of $\mathcal{T}$ have the same cardinality).
Hence, probability densities of input point patterns relative to $\mu$
are not indicative of how well they fit the 'normal' data model. 

Remark: In novelty detection we are interested in the likelihood of
the input point pattern whereas in Bayesian classification we are
interested in its likelihood ratio. The posterior class probability
\begin{align*}
p\left(y\mid X\right) & =\frac{p(y)f(X\mid y)}{\int p(y)f(X\mid y)dy}\\
 & =\frac{p(y)P(dX\mid y)/\mu(dX)}{\int p(y)(P(dX\mid y)/\mu(dX))dy}\\
 & =\frac{p(y)P(dX\mid y)}{\int p(y)P(dX\mid y)dy},
\end{align*}
(using standard properties of Radon-Nikodym derivative and relevant
absolute continuity assumption) is the ratio, at an infinitesimal
neighbourhood $dX$, between the joint probability $P(dX,y)$, and
the probability $P(dX)$, which is invariant to the choice of reference
measure. In essence, the normalizing constant cancels out the influence
of the reference measure, and hence, problems with the non-uniformity
of the reference measure do not arise. 

\subsection{Ranking functions\label{sec:ranking-function}}

To the best of our knowledge, it is not known whether there exists
a uniform reference measure on $\mathcal{F}(\mathcal{X})$ that dominates
the probability distributions of interest (so that they admit densities).
In this subsection, we propose a suitable point pattern ranking function
for novelty detection by modifying the probability density. 

The probability density (\ref{eq:generalRFSdensity}) is the product
of the cardinality distribution $p_{c}(|X|)$, the cardinality-conditioned
feature (probability) density $f_{|X|}(X)$, and a trans-dimensional
weight $|X|!U^{|X|}$. Note that the cardinality distribution and
the conditional joint feature density completely describes the point
process. The conditional density $f_{|X|}(X)$ enables the ranking
of point patterns of the same cardinality, but cannot be used to rank
across different cardinalities because it takes on different units
of measurement. The weights $|X|!U^{|X|}$ reconcile for the differences
in dimensionality and unit of measurement between $f_{|X|}(X)$ of
different cardinalities. However, the example in subsection \ref{sec:PDF-likelihood}
demonstrates that weighting by $|X|!U^{|X|}$ leads to probability
densities that are inconsistent with likelihoods. 

In the generalization of the maximum aposteriori (MAP) estimator to
point patterns \cite{mahler2007statistical}, Mahler circumvented
such inconsistency by replacing $|X|!U^{|X|}$ with $c^{|X|}$ , where
$c$ is an arbitrary constant. More concisely, instead of maximizing
the probability density $f(X)$, Mahler proposed to maximize $f(X)c^{|X|}/|X|!$.
Since $c$ is a free parameter, the generalized MAP estimate depends
on the choice of $c$. 

Inspired by Mahler's generalized MAP estimator, we replace the weight
$|X|!U^{|X|}$ in the probability density by a general function of
the cardinality $C(|X|)$, resulting in a ranking function of the
form\vspace{-1mm}
\begin{equation}
r(X)=p_{c}(|X|)C(|X|)f_{|X|}(X).\label{eq:new_RFSlikeli_with_c}
\end{equation}
The example in subsection \ref{sec:PDF-likelihood} demonstrated that,
as a function of cardinality, the ranking should be proportional to
the cardinality distribution, otherwise unlikely samples can assume
high ranking values. In general, the ranking function is not solely
dependent on the cardinality, but also varies with the features. Nonetheless,
the example suggests that the ranking function, on average, should
be proportional to the cardinality distribution. Hence, we impose
the following consistency requirement: for a given cardinality $n$,
the expected ranking value is proportional to the probability of cardinality
$n$, i.e.,\vspace{-1mm}
\begin{equation}
\mathbb{E}_{X\mid|X|=n}\left[r(X)\right]\propto p_{c}(n).\label{eq:conditional}
\end{equation}

\begin{prop}
\label{prop:ranking}For a point process with probability density
(\ref{eq:generalRFSdensity}), a ranking function consistent with
the cardinality distribution, i.e., satisfies (\ref{eq:conditional}),
is given by\vspace{-2mm}
\begin{equation}
r(X)\propto\,\frac{p_{c}(|X|)}{||f_{|X|}||_{2}^{2}}f_{|X|}(X)\label{eq:new_RFSlikeli_final-1-1}
\end{equation}
where $||\cdot||_{2}$ denotes the $L_{2}$-norm. 
\end{prop}
\begin{IEEEproof}
Noting $f(X\negthinspace\mid\negthinspace|X|=n)=n!U^{n}f_{n}(X)\delta_{n}[|X|]$
from (\ref{eq:generalRFSdensity}), and using the integral (\ref{eq:commonRefMeasure-1})
we have
\begin{align*}
\mathbb{\mathbb{E}}_{X\mid|X|=n}\left[f_{n}(X)\right] & =\int f_{n}(X)\,f(X\negthinspace\mid\negthinspace|X|=n)\,\mu(dX)\\
 & =\frac{n!U^{n}}{n!U^{n}}\int(f_{n}(\{x_{1},...,x_{n}\}))^{2}d(x_{1},...,x_{n})\\
 & =||f_{n}||_{2}^{2}.
\end{align*}
Hence\vspace{-2mm}
\begin{align*}
\mathbb{E}_{X\mid|X|=n}\left[r(X)\right] & \propto\mathbb{\mathbb{E}}_{X\mid|X|=n}\left[\frac{p_{c}(|X|)}{||f_{|X|}||_{2}^{2}}f_{|X|}(X)\right]\\
 & =\frac{p_{c}(n)}{||f_{n}||_{2}^{2}}\mathbb{\mathbb{E}}_{X\mid|X|=n}\left[f_{n}(X)\right]\\
 & =p_{c}(n).
\end{align*}

\vspace{-6mm}
\end{IEEEproof}
Note that $||f_{|X|}||_{2}^{2}$ has units of $U^{-|X|}$ , which
is the same as the unit of $f(X)$, rendering the ranking function
$r$ unit-less, thereby avoids the unit of measurement inconsistency
described in Section \ref{subsec:Introduction}.

\begin{figure}[h]
\begin{centering}
\vspace{-3mm}
\subfloat[\label{fig:Gau_L2norm-fat}]{\centering{}\includegraphics[width=0.48\columnwidth]{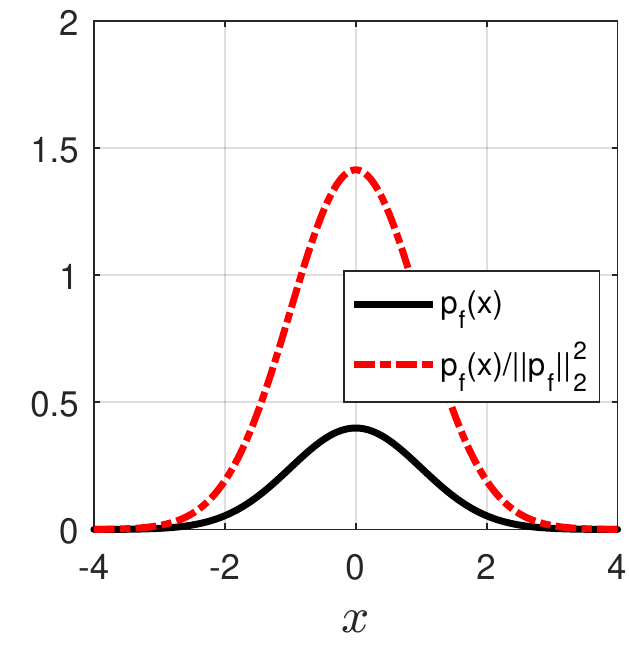}}~\subfloat[\label{fig:Gau_L2norm-thin}]{\centering{}\includegraphics[width=0.48\columnwidth]{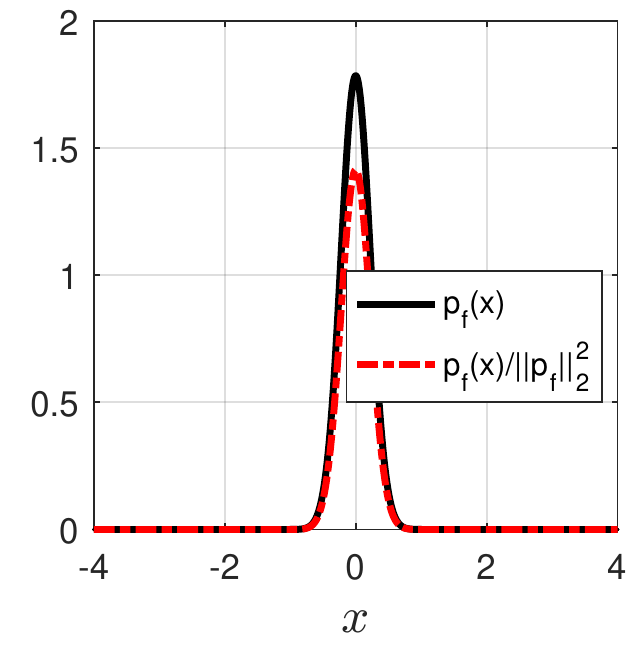}}
\par\end{centering}
\caption{\label{fig:Gau_L2norm}Probability density divided by energy: (a)
`short' Gaussian (mean = 0, variance = 1); (b) `tall' Gaussian (mean
= 0, variance = 0.05). }
\vspace{-2mm}
\end{figure}

For an IID-cluster with feature density $p_{f}$ the ranking function
reduces to\vspace{-1mm}
\begin{equation}
r(X)\propto p_{c}(|X|)\left(\frac{p_{f}}{||p_{f}||_{2}^{2}}\right)^{X}.\label{eq:new_RFSlikeli_final-1-1-1}
\end{equation}
The feature density $p_{f}$, in the example of subsection \ref{sec:PDF-likelihood},
is uniform and so $p_{f}/||p_{f}||_{2}^{2}=1$ on its support. Hence
the ranking is equal to the cardinality distribution, as expected.
Fig. \ref{fig:Gau_L2norm} illustrates the effect of dividing a non-uniform
feature density $p_{f}$, by its energy $||p_{f}||_{2}^{2}$: `tall'
densities become shorter and `short' densities become taller, providing
adjustments for multiplying together many large/small numbers.  

\section{Model-Based Clustering\label{sec:Model-Based-MI-Clustering}}

The aim of clustering is to partition the dataset into groups so that
members in a group are similar to each other whilst dissimilar to
observations from other groups~\cite{murphy2012machine}. A partitioning
of a given set of observations $\{X_{1}$,...,$X_{N}$\} is often
represented by the (latent) cluster assignment $y_{1:N}$, where $y_{n}$
denotes the cluster label for the $n$\textsuperscript{th} observation.
Clustering is an unsupervised learning problem since the labels are
not included in the observations \cite{Jain1999data_clustering,russell2003artificial}.
Indeed it can be regarded as classification without training and is
a fundamental problem in data analysis. Comprehensive surveys on clustering
can be found in \cite{Jain1999data_clustering,jain2010clustering50yearsKmeans,xu2005survey_clustering}. 

At present, model-based point pattern clustering have not been investigated.
To the best of our knowledge, there are two clustering algorithms
for point patterns: the Bag-level MI Clustering (BAMIC) algorithm
\cite{zhang2009MIClustering}; and the Maximum Margin MI Clustering
(M$^{3}$IC) algorithm \cite{zhang2009m3icClustering}. BAMIC adapts
the $k$-medoids algorithm with the Hausdorff distance as a measure
of dissimilarity between point patterns \cite{zhang2009MIClustering}.
On the other hand, in M$^{3}$IC, the clustering was posed as a non-convex
optimization problem which is then relaxed and solved via a combination
of the Constrained Concave-Convex Procedure and Cutting Plane methods
\cite{zhang2009m3icClustering}. While these algorithms are simple
to use, they lack the ability to exploit statistical trends in the
data, not to mention computational problems with high dimensional
or large datasets \cite{murphy2012machine}. 

In this section, we propose a model-based approach to the clustering
problem for point pattern data. Mixture modeling is the most common
probabilistic approach to clustering, where the aim is to estimate
the cluster assignment $y_{1:N}$ via likelihood or posterior inference
\cite{murphy2012machine}. The point process formalism enables direct
extension of mixture models to point pattern data. In particular,
the finite mixture point process model for problems with known number
of clusters is presented in subsection \ref{subsec:Mixture-of-RFS},
while the infinite mixture point process model for problems with unknown
number of clusters is presented in subsection \ref{subsec:infinite-mixture}.

\subsection{Finite Mixture Model \label{subsec:Mixture-of-RFS}}

A finite mixture model assumes $K$ underlying clusters labeled $1$
to $K$, with prior probabilities $\pi_{1},...,\pi_{K}$, and characterized
by the parameters $\theta_{1},...,\theta_{K}$ in some space $\Theta$.
Let $f(X_{n}\negthinspace\mid\negthinspace\theta_{k})\triangleq f(X_{n}\negthinspace\mid\negthinspace y_{n}=k,\theta_{1:K})$
denote the likelihood of $X_{n}$ given that cluster $k$ generates
an observation. Then
\begin{equation}
f(X_{1:N},y_{1:N}\negthinspace\mid\negthinspace\pi_{1:K},\theta_{1:K})=\prod_{n=1}^{N}\pi_{y_{n}}\,f(X_{n}\negthinspace\mid\negthinspace\theta_{y_{n}}),\label{eq:mixture-model}
\end{equation}
Marginalizing the joint distribution (\ref{eq:mixture-model}) over
the cluster assignment $y_{1:N}$ gives the data likelihood function
\begin{equation}
f(X_{1:N}\negthinspace\mid\negthinspace\pi_{1:K},\theta_{1:K})=\prod_{n=1}^{N}\sum_{k=1}^{K}\pi_{k}\,f(X_{n}\negthinspace\mid\negthinspace\theta_{k}).\label{eq:mixture-of-RFSs}
\end{equation}
Thus, in a finite mixture model, the likelihood of an observation
is a mixture of $K$ probability densities. Hence the application
of the finite mixture model requires the number of clusters to be
known apriori. The posterior probability of cluster label $y_{n}=k$
(i.e., the probability that, given $\pi_{1:K},\theta_{1:K}$ and $X_{n}$,
cluster $k$ generates $X_{n}$) is 
\begin{equation}
p\left(y_{n}=k\negthinspace\mid\negthinspace X_{n},\pi_{1:K},\theta_{1:K}\right)=\frac{\pi_{k}\,f(X_{n}\negthinspace\mid\negthinspace\theta_{k})}{\sum_{\ell=1}^{K}{\displaystyle \pi_{\ell}\,f(X_{n}\negthinspace\mid\negthinspace\theta_{\ell})}}.\label{eq:cluster-posterior}
\end{equation}

Under a mixture model formulation, clustering can be treated as an
incomplete data problem since only the $X_{1:N}$ of the complete
data $\mathcal{D}=\{(X_{n},y_{n})\}_{n=1}^{N}$ is observed and the
cluster assignment $y_{1:N}$ is unknown or missing. We seek $y_{1:N}$,
and the mixture model parameter $\psi\triangleq(\pi_{1:K},\theta_{1:K})$
that best explain the observed data $X_{1:N}$ according to a given
criterion such as ML or optimal Bayes. ML is intractable in general
and often requires the Expectation-Maximisation (EM) algorithm \cite{dempster1977em_algorithm,Little2002StatisMissingData_EM}
to find approximate solutions. Optimal Bayes requires suitable priors
for $\psi$. Typically the prior for $\pi_{1:K}$ is a Dirichlet distribution
$Dir(\cdot|\eta/K,...,\eta/K)$ with dispersion $\eta$, while the
prior for $\theta_{1:K}$ is model-specific, depending on the form
of the likelihood $f(X_{n}|\theta_{k})$. Computing the cluster label
posterior $p(y_{1:N}|X_{1:N})$ or the joint posterior $p(y_{1:N},\psi|X_{1:N})$
are intractable in general and Markov Chain Monte Carlo methods, such
as Gibbs sampling are often needed \cite{bernardo2009bayesian,Gelman_etal_bk03_Bayesian}.

Next we detail an ML solution to point pattern clustering using EM
with an IID-cluster mixture model. Instead of presenting a Bayesian
solution for the finite mixture model, in subsection \ref{subsec:infinite-mixture}
we extend it to address unknown number of clusters, and develop a
solution based on Gibbs sampling, which can be simplified to the finite
mixture case.

\subsubsection{EM clustering via IID-cluster mixture model\label{subsec:EM-clustering}}

The EM algorithm maximizes the data likelihood (\ref{eq:mixture-of-RFSs})
by generating a sequence of iterates $\{\psi^{(i)}\}_{i=0}^{\infty}$
using the following two steps \cite{dempster1977em_algorithm,Little2002StatisMissingData_EM}:
\begin{itemize}
\item \emph{E-step:} Compute $Q(\psi\negthinspace\mid\negthinspace\psi^{(i-1)})$,
defined as
\begin{align*}
 & \negthinspace\negthinspace\negthinspace\negthinspace\mathbb{E}_{y_{1:N}\negthinspace\mid\negthinspace X_{1:N},\psi^{(i-1)\negthinspace}}\left[\log\negthinspace f(X_{1:N\negthinspace},y_{1:N}\negthinspace\mid\negthinspace\psi)\right]\\
\; & \;\;\;=\sum_{k=1}^{K}\sum_{n=1}^{N}\negthinspace\log\negthinspace\left(\pi_{k}f(X_{n\negthinspace}\negthinspace\mid\negthinspace\theta_{k\negthinspace})\right)\negthinspace p(y_{n}\negthinspace=\negthinspace k\negthinspace\mid\negthinspace X_{n},\psi^{(i-1)}).
\end{align*}
\item \emph{M-step:} Find $\psi^{(i)}=\underset{\psi}{\text{argmax\,}}Q(\psi\negthinspace\mid\negthinspace\psi^{(i-1)})$
.
\end{itemize}
The expectation $Q(\psi^{(i)}\negthinspace\mid\negthinspace\psi^{(i-1)})$
increases after each EM iteration, and consequently converges to a
(local) maximum of (\ref{eq:mixture-of-RFSs}) \cite{dempster1977em_algorithm,Little2002StatisMissingData_EM}.
In practice, the iteration is terminated at a user defined number
$N_{\mathrm{iter}}$ or when increments in $Q(\psi^{(i)}\negthinspace\mid\negthinspace\psi^{(i-1)})$
falls below a given threshold.%
\begin{comment}
FOR THESIS: 

1. For setting the number of cluster see \cite{dasgupta2000two_round_EM}

2. Show the log version of EM (log-sum-exp trick).

3. Discuss use Cauchy Swart divergence to stop EM.
\end{comment}
{} The optimal cluster label estimate is the mode of the cluster label
posterior (\ref{eq:cluster-posterior}). 

Following the arguments from \cite{bilmes1998gentle_EM}, the M-step
can be accomplished by separately maximizing $Q(\pi_{1:K},\theta_{1:K}\negthinspace\mid\negthinspace\psi^{(i-1)})$
over $\theta_{1},...,\theta_{K}$ and $\pi_{1:K}$. Using Lagrange
multiplier with constraint $\sum_{k=1}^{K}\pi_{k}\negthinspace=\negthinspace1$,
yields the optimal weights:
\begin{equation}
\pi_{k}^{(i)}=\frac{1}{N}\sum_{n=1}^{N}p(y_{n}=k\negthinspace\mid\negthinspace X_{n},\psi^{(i\mbox{-}1)}).\label{eq:comp_weight}
\end{equation}
Noting that $\log\negthinspace\left(\pi_{k}f(X_{n\negthinspace}\negthinspace\mid\negthinspace\theta_{k\negthinspace})\right)$
is accompanied by the weight $p(y_{n}\negthinspace=\negthinspace k\negthinspace\mid\negthinspace X_{n},\psi^{(i-1)})$,
\textit{maximizing} $Q(\pi_{1:K},\theta_{1:K}\negthinspace\mid\negthinspace\psi^{(i-1)})$
\textit{over} $\theta_{k}$ \textit{is equivalent to ML estimation
of} $\theta_{k}$ \textit{with weighted data}. However, the data-weighted
MLE of $\theta_{k}$ depends on the specific form of $f(\cdot|\theta_{k})$,
and is intractable in general. 

Fortunately, for the IID-cluster mixture model, where
\begin{equation}
f(X\negthinspace\mid\negthinspace\theta_{k})=p_{\xi_{k}}(|X|)\,|X|!\,U^{|X|}p_{\varphi_{k}}^{X},\label{eq:component-density}
\end{equation}
with $\theta_{k}=(\xi_{k},\varphi_{k})$ denoting the parameters of
the cardinality and feature distributions, tractable solutions are
available. Similar to Proposition \ref{prop:IID-cluster},\textit{
the IID-cluster form allows the data-weighted MLE of} $\theta_{k}$
\textit{to separate into data-weighted MLEs of} $\xi_{k}$ \textit{and}
$\varphi_{k}$. Some examples are: 
\begin{itemize}
\item For a categorical cardinality distribution with maximum cardinality
$M$, where $\xi_{k}=(\xi_{k,0},...,\xi_{k,M})$ lies in the unit
$M$-simplex, the iteration is
\[
\xi_{k,m}^{(i)}=\frac{\sum_{n=1}^{N}\delta_{m}\,[|X_{n}|]p(y_{n}=k\negthinspace\mid\negthinspace X_{n},\psi^{(i\mbox{-}1)})}{\sum_{\ell=0}^{M}\sum_{n=1}^{N}\delta_{\ell}[|X_{n}|]\,p(y_{n}=k\negthinspace\mid\negthinspace X_{n},\psi^{(i\mbox{-}1)})};
\]
\item For a Poisson cardinality distribution, where $\xi_{k}>0$ is the
mean cardinality, the iteration is
\[
\xi_{k}^{(i)}=\frac{\sum_{n=1}^{N}|X_{n}|\,p(y_{n}=k\negthinspace\mid\negthinspace X_{n},\psi^{(i\mbox{-}1)})}{\sum_{n=1}^{N}p(y_{n}=k\negthinspace\mid\negthinspace X_{n},\psi^{(i\mbox{-}1)})};
\]
\item For a Gaussian feature distribution, where $\varphi_{k}=(\mu_{k},\Sigma_{k})$
is the mean-covariance pair, the iteration is
\begin{align*}
\mu_{k}^{(i)}= & \frac{\sum_{n=1}^{N}p(y_{n}=k\negthinspace\mid\negthinspace X_{n},\psi^{(i\mbox{-}1)})\,\sum_{x\in X_{n}}x}{\sum_{n=1}^{N}{\displaystyle |X_{n}|}\,p(y_{n}=k\negthinspace\mid\negthinspace X_{n},\psi^{(i\mbox{-}1)})},\\
\Sigma_{k}^{(i)}= & \frac{\sum_{n=1}^{N}{\displaystyle p(y_{n}=k\negthinspace\mid\negthinspace X_{n},\psi^{(i\mbox{-}1)})}\sum_{x\in X_{n}}K_{k}^{(i)}(x)}{\sum_{n=1}^{N}{\displaystyle |X_{n}|\,p(y_{n}=k\negthinspace\mid\negthinspace X_{n},\psi^{(i\mbox{-}1)})}},
\end{align*}
where $K_{k}^{(i)}(x)=(x-\mu_{k}^{(i)})(x-\mu_{k}^{(i)})^{\mathrm{T}}$;
\item For a Gaussian mixture feature distribution, where $\varphi_{k}$
is the Gaussian mixture parameter, $\varphi_{k}^{(i)}$ can be determined
by applying the standard EM algorithm on the weighted data.
\end{itemize}

\subsection{Infinite Mixture Model\label{subsec:infinite-mixture}}

For an unknown number of clusters, finite mixture models are no longer
directly applicable. Bayesian non-parametric modeling (see e.g., \cite{Ghosh_bk03_Bayesian,Hjort_etal_bk10_bayesian})
addresses the unknown number of clusters by modeling the set of mixture
parameters as a point process. Thus, the observations and the clusters
are all modeled as point processes.

In a finite mixture model, the number of components (and clusters)
is fixed at $K$. The mixture parameter $\psi=(\pi_{1:K},\theta_{1:K})$
is a point in $(\mathbb{R}_{+}\times\Theta)^{K}$, such that $\sum_{i}^{K}\pi_{i}=1$.
Under the Bayesian framework, it is further assumed that $\theta_{1:K}$
follows a given distribution on $\Theta^{K}$, and that $\pi_{1:K}$
follows a distribution on the unit $(K-1)$-simplex, e.g. a Dirichlet
distribution. 

An infinite mixture model addresses the unknown number of components
by considering the mixture parameter $\varPsi$ as a point pattern
in $\mathbb{R}_{+}\times\Theta$ such that $\sum_{(\pi,\theta)\in\varPsi}\pi=1$.
Further, under the Bayesian non-parametric framework, we furnish $\varPsi$
with a prior distribution, thereby modeling the mixture parameter
as a point process on $\mathbb{R}_{+}\times\Theta$. The simplest
model would be the Poisson point process, but the resulting component
weights do not necessarily sum to one. Nonetheless, these weights
can be normalized to yield a tractable point process model for the
mixture parameter \cite{Lin_etal_10construction,Jordan_10hierarchical}.
More concisely, let $\Xi$ be a Poisson point process on $\mathbb{R}_{+}\times\Theta$
with intensity measure $\eta\omega^{-1}e^{-\eta\omega}d\omega G_{0}(d\theta)$,
i.e., the product of an improper gamma distribution and the base distribution
$G_{0}$. Then the prior model for the mixture parameter is given
by
\begin{equation}
\varPsi=\left\{ \left(\nu_{\Xi}^{-1}\omega,\theta\right):(\omega,\theta)\in\Xi\right\} ,\label{eq:Poisson-Dirichlet}
\end{equation}
where $\nu_{\Xi}=\sum_{(\omega,\theta)\in\Xi}\omega$. Note that (\ref{eq:Poisson-Dirichlet})
is no longer a Poisson point process because each constituent element
involves the sum $\nu_{\Xi}$, thereby violating the independence
condition. 

To specify an explicit form for the prior distribution of the mixture
parameter $\varPsi$, note that each point $\left(\nu_{\Xi}^{-1}\omega,\theta\right)$
can be equivalently represented by atom at $\theta$ with weight $\nu_{\Xi}^{-1}\omega$,
and hence the point process (\ref{eq:Poisson-Dirichlet}) can be represented
by the random atomic distribution $G$ on $\Theta$, defined by

\begin{equation}
G(A)=\nu_{\Xi}^{-1}\sum_{(\omega,\theta)\in\Xi}\omega\,\mathbf{1}_{A}(\theta).
\end{equation}
It was shown in \cite{Lin_etal_10construction} that $G$ follows
a Dirichlet process $DP(\eta,G_{0})$, with parameter $\eta$ and
base distribution $G_{0}$. Noting that the cluster parameter $\theta_{n}$
for $X_{n}$ can be regarded as a sample from $G$, the data generation
process for this model can be summarized as follows
\begin{align*}
G & \sim DP(\eta,G_{0})\\
\theta_{n} & \sim G\\
X_{n} & \sim f(\cdot|\theta_{n}).
\end{align*}

The cluster assignment and the mixture parameters, including the number
of clusters, can be automatically learned from the data via posterior
inference. Analogous to finite mixture models, computing the posteriors
are intractable in general and often require MCMC methods. Next we
detail a (collapsed) Gibbs sampler to simulate the cluster label posterior
$p(y_{1:N}|X_{1:N})$ for point pattern clustering. 

\subsubsection{Gibbs Sampling for Poisson mixture model\label{subsec:EM-clustering-1}}

A Gibbs sampler simulates the posterior $p(y_{1:N}|X_{1:N})$ via
a Markov Chain $\{y_{1:N}^{(i)}\}_{i=0}^{\infty}$, in which $y_{1:N}^{(i)}$
is generated from $y_{1:N}^{(i-1)}$ by iterating:
\[
y_{n}^{(i)}\sim p_{n}(\cdot|y_{1:n\negthinspace-\negthinspace1}^{(i)},y_{n+1:N\negthinspace}^{(i-1)},X_{1:N\negthinspace}),
\]
for $n=1,...,N$, where
\[
p_{n}(y_{n}|y_{1:n\negthinspace-\negthinspace1\negthinspace},y_{n+1:N\negthinspace},X_{1:N\negthinspace})\propto p(y_{1:N\negthinspace}|X_{1:N\negthinspace})
\]
is the $n$\textsuperscript{th} conditional probability. After the
so-called pre-convergence period, samples from the Markov chain are
effectively distributed from the cluster label posterior. However,
the actual distribution of the samples depends on the starting value.
In practice, the pre-convergence samples, known as burn-ins, are discarded,
and the post-convergence samples are used for inference. 

While Gibbs sampling is efficient, it requires the conditionals to
be easily computed and sampled. Using the notation $\bar{n}\triangleq\{1,...,N\}-\{n\}$,
the $n$\textsuperscript{th} conditional for the cluster labels of
the infinite point process mixture can be written as
\begin{align}
\negthinspace\negthinspace\negthinspace\negthinspace p(k|y_{\bar{n}}, & X_{1:N})=p(y_{n}=k|z_{\bar{n}},X_{n},X_{\bar{n}})\nonumber \\
\negthinspace\negthinspace\negthinspace\negthinspace\negthinspace\negthinspace\negthinspace\negthinspace\propto & f(X_{n}|y_{n}=k,y_{\bar{n},},X_{\bar{n}})p(y_{n}=k|y_{\bar{n}},X_{\bar{n}})\nonumber \\
\negthinspace\negthinspace\negthinspace\negthinspace\negthinspace\negthinspace\negthinspace\negthinspace= & f(X_{n}|y_{n}=k,y_{\bar{n},},X_{\bar{n}})p(y_{n}=k|y_{\bar{n}})\label{eq:CL-conditionals}
\end{align}
where the last line follows from the fact that given $X_{\bar{n}}$
and $y_{\bar{n}}$, $y_{n}$ is independent of $X_{\bar{n}}$. Using
the Polya urn characterization \cite{Blackwell_MacQueen_73ferguson}
of the Dirichlet process, we have
\begin{align}
p(y_{n}=k|y_{\bar{n}}) & =\frac{(1\negthinspace-\negthinspace\mathbf{1}_{\{y_{\bar{n}}\}\negthinspace}(k))\eta+\negthinspace\sum_{j\in\bar{n}\negthinspace}\delta_{y_{j\negthinspace}}[k]}{N-1+\eta}\label{eq:component-density-1-1}
\end{align}
where the sum over $\bar{n}$ is called the \textit{popularity} of
$k$. Further, note in the first term of (\ref{eq:CL-conditionals})
that, given $y_{n}=k,$ $y_{\bar{n}}$ and $X_{\bar{n}}$, $X_{n}$
only depends on $X_{\bar{n}}^{(k)}\triangleq\{X_{j}\in X_{\bar{n}}:y_{j}=k\}$,
the set of observations in $X_{\bar{n}}$ that belongs to cluster
$k$ (in fact its cardinality is the popularity of $k$). Hence
\begin{align}
f(X_{n}|y_{n}=k,y_{\bar{n},},X_{\bar{n}}) & =f(X_{n}|y_{n}=k,X_{\bar{n}}^{(k)})\nonumber \\
 & =\int\negthinspace f(X_{n}|\theta_{k})G(d\theta_{k}|X_{\bar{n}}^{(k)}),\label{eq:predictivelikelihood}
\end{align}
which is the predictive likelihood for cluster $k$ of the point pattern
$X_{n}$ given $X_{\bar{n}}^{(k)}$. In general the predictive likelihood,
and consequently the conditionals, are intractable. 

Fortunately, an analytic predictive likelihood is available for the
(infinite) Poisson point process mixture, where 
\begin{equation}
f(X\negthinspace\mid\negthinspace\theta_{k})=\xi_{k}^{|X|}\,e^{-\xi_{k}}U^{|X|}p_{\varphi_{k}}^{X}\,\label{eq:component-density-1}
\end{equation}
with $\theta_{k}=(\xi_{k},\varphi_{k})$ denoting the mean cardinality
and feature density parameter. In the following, we propose a family
of conjugate priors for Poisson point processes and exploit conjugacy
to derive the predictive likelihood for the Poisson mixture model. 
\begin{prop}
\label{prop:conjugate}Let $H(d\varphi|\gamma)$ be a conjugate prior
distribution with respect to the feature density $p_{\varphi}$ of
a Poisson point process likelihood $f(X\negthinspace\mid\negthinspace\xi,\varphi)$.
Then the distribution given by
\begin{equation}
G_{0}(d\xi,d\varphi|\alpha,\beta,\gamma)=\frac{\beta^{\alpha}}{\Gamma(\alpha)}\xi^{\alpha-1}e^{-\beta\xi}d\xi H(d\varphi|\gamma)\label{eq:conjugate}
\end{equation}
is conjugate with respect to $f(X\negthinspace\mid\negthinspace\xi,\varphi)$.
Moreover, the predictive likelihood of $X$ given a finite collection
$\mathcal{Z}$ of point patterns is
\begin{align}
f(X|\mathcal{Z})= & \frac{\Gamma(\alpha{}_{_{\mathcal{Z}}}+|X|)\beta_{_{\mathcal{Z}}}^{\alpha{}_{_{\mathcal{Z}}}}}{\Gamma(\alpha{}_{_{\mathcal{Z}}})(\beta{}_{_{\mathcal{Z}}}+1)^{\alpha{}_{_{\mathcal{Z}}}+|X|}}\int\negthinspace p_{\varphi}^{X}H(d\varphi|\gamma_{_{\mathcal{Z}}}).\label{eq:PoissonPredictiveLikelihood}
\end{align}
where $\alpha{}_{_{\mathcal{Z}}}=\alpha+\sum_{Z\in\mathcal{Z}}|Z|$,
$\beta{}_{_{\mathcal{Z}}}=\beta+|\mathcal{Z}|$ and $H(d\varphi|\gamma_{_{\mathcal{Z}}})\propto\prod_{Z\in\mathcal{Z}}p_{\varphi}^{Z}H(d\varphi|\gamma)$. 
\end{prop}
\begin{IEEEproof}
Since $H(d\varphi|\gamma)$ and $p_{\varphi}$ are conjugate, i.e.,
$H(d\varphi|\gamma)$ and $p_{\varphi}(z)H(d\varphi|\gamma)/\negthinspace\int\negthinspace p_{\varphi}(z)H(d\varphi|\gamma)$
have the same form, it follows by induction that $p_{\varphi}^{Z}H(d\varphi|\gamma)/\negthinspace\int\negthinspace p_{\varphi}^{Z}H(d\varphi|\gamma)$
also has the same form, which we denote by $H(d\varphi|\gamma_{_{Z}})$.
Using Bayes\textquoteright{} rule 
\begin{align*}
G(d\xi,d\varphi|Z,\alpha,\beta,\gamma) & \propto f(Z\negthinspace\mid\negthinspace\xi,\varphi)G_{0}(d\xi,d\varphi|\gamma,\alpha,\beta)\\
 & \propto\xi^{|Z|}\,e^{-\xi}U^{|Z|}p_{\varphi}^{Z}\xi^{\alpha-1}e^{-\beta\xi}d\xi H(d\varphi|\gamma)\\
 & \propto\xi^{|Z|+\alpha-1}\,e^{-(\beta+1)\xi}d\xi H(d\varphi|\gamma_{_{Z}})
\end{align*}
which takes on the same form as (\ref{eq:conjugate}). Hence (\ref{eq:conjugate})
is conjugate with respect to $f(Z\negthinspace\mid\negthinspace\xi,\varphi)$.
Further, iterating the above argument through the elements of $\mathcal{Z}$
yields
\[
G(d\xi,d\varphi|\mathcal{Z},\alpha,\beta,\gamma)\propto\xi^{\alpha{}_{_{\mathcal{Z}}}-\negthinspace1}e^{-\beta{}_{_{\mathcal{Z}}}\xi}d\xi H(d\varphi|\gamma_{_{\mathcal{Z}}})
\]
Substituting this into the predictive likelihood
\begin{align*}
f(X|\mathcal{Z})= & \int\negthinspace\negthinspace\int\negthinspace f(X|\xi,\varphi)G(d\xi,d\varphi|\mathcal{Z},\alpha,\beta,\gamma)
\end{align*}
together with some algebraic manipulations yields (\ref{eq:PoissonPredictiveLikelihood}).
\end{IEEEproof}
For an infinite Poisson mixture model with base measure $G_{0}$ given
by (\ref{eq:conjugate}), where $H(d\varphi|\gamma)$ is conjugate
with respect to the feature density $p_{\varphi}$ of the constituent
Poisson point process components, the predictive likelihood (\ref{eq:predictivelikelihood})
is given by $f(X_{n}|X_{\bar{n}}^{(k)})$. Depending on the specific
forms for $H(d\varphi|\gamma)$ and $p_{\varphi}$, the integral $\int\negthinspace p_{\varphi}^{X}H(d\varphi|\gamma_{_{\mathcal{Z}}})$
can be evaluated analytically (see \cite{bernardo2009bayesian} for
several examples). Consequently, given the hyper-parameters $\alpha,\beta,\gamma,\eta$
the conditionals for the Gibbs sampler can be computed analytically.

Remark: The proposed Bayesian solution can be adapted for semi-supervised
learning, where only labeled training data for certain clusters are
available and the objective is to compute the posterior of the missing
labels. This approach can also address the novelty detection problem
in Section \ref{sec:Anomaly-detection} without having to rank the
input observations, albeit at greater computational cost.

\section{Experiments\label{sec:Experiments}}

This section demonstrates the viability of the proposed framework
with one of the simplest point processes\textendash the Poisson model.
A Poisson point process with Gaussian intensity is specified by the
triple $(\rho,\mu,\Sigma)$ where $\rho$ is the rate and $\mu,\Sigma$
are the mean and covariance of the feature density. The NB model is
used as a performance bench mark since it has been used for this type
of problems (see e.g. \cite{maron1961NB_setSizeVaried,joachims1996probabilistic,mccallum1998comparison_NBtextClassifi,csurka2004visual,cadez2000EMclustering_VariableLengthData})
and assumes i.i.d. features like the Poisson model.\vspace{-0mm}

\subsection{Classification Experiments}

This subsection presents two classification experiments on simulated
data and the Texture images dataset \cite{textureDataset}. In the
training phase, ML is used to learn the parameters of the NB model
and the Poisson model (using the technique outlined in subsection
\ref{subsec:ML-Learning-IID}) from fully observed training data.
For simplicity we use a uniform class prior in the test phase.\vspace{-0mm}

\subsubsection{Classification on simulated data\label{subsec:sim-classifi}}

\begin{figure*}[t]
\begin{centering}
\vspace{-0mm}
\subfloat[\label{fig:Classifi_Sim_sepa_feat}All 3 classes are well-separated
from each other in feature, but overlapping in cardinality. ]{\begin{centering}
\includegraphics[bb=0bp 20mm 163bp 220bp,height=4.25cm]{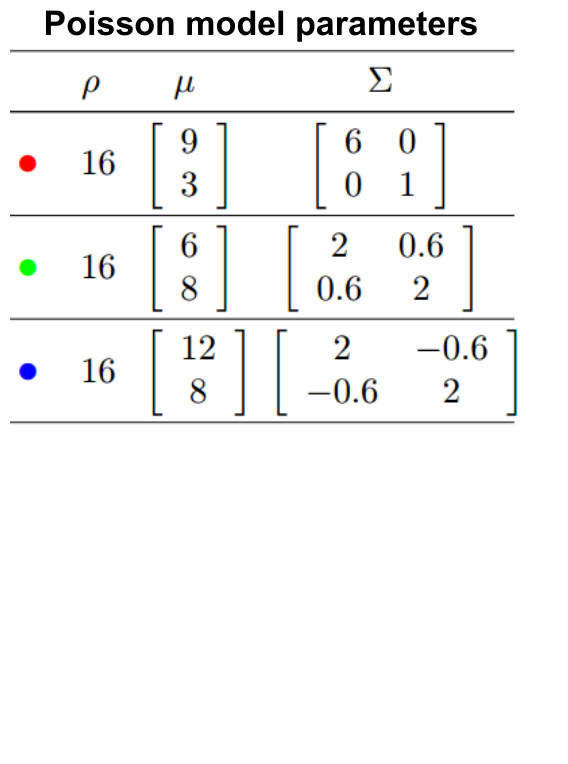}\includegraphics[height=4.2cm]{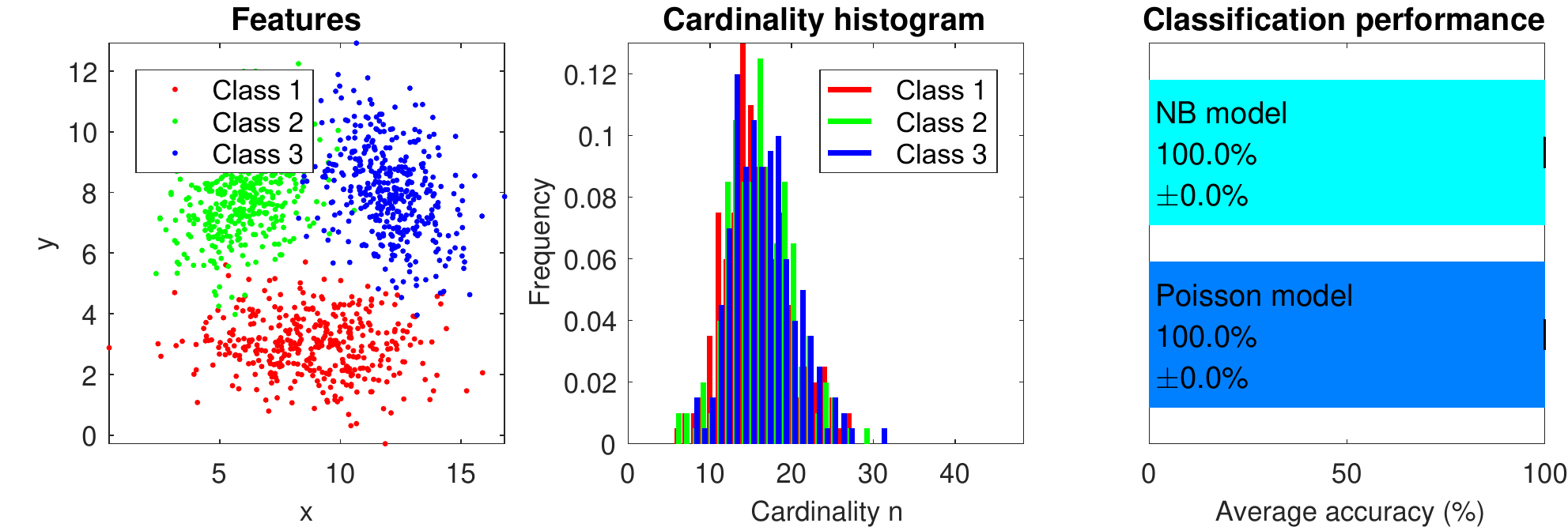}
\par\end{centering}
\centering{}\vspace{1mm}
}
\par\end{centering}
\begin{centering}
\vspace{-2mm}
\subfloat[\label{fig:Classifi_Sim_overlap_feat}All 3 classes are overlapping
in feature, but well-separated from each other in cardinality. ]{\begin{centering}
\includegraphics[bb=0bp 20mm 163bp 220bp,height=4.25cm]{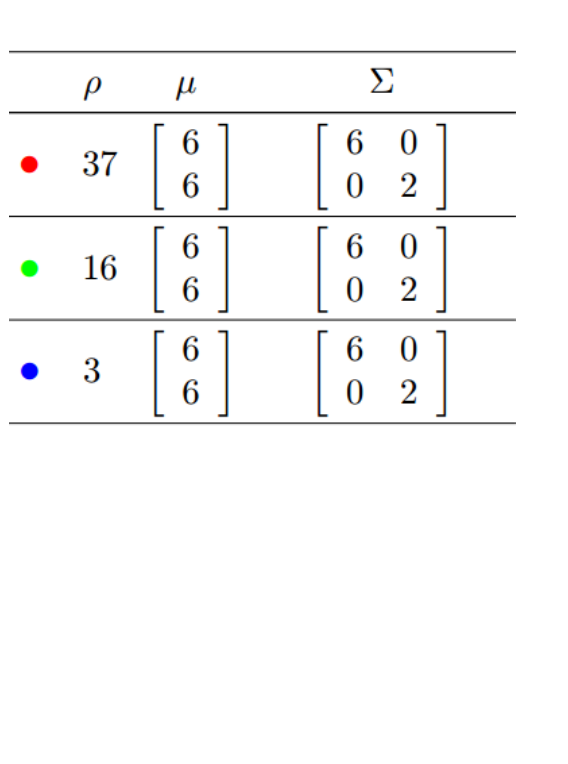}\includegraphics[height=4cm]{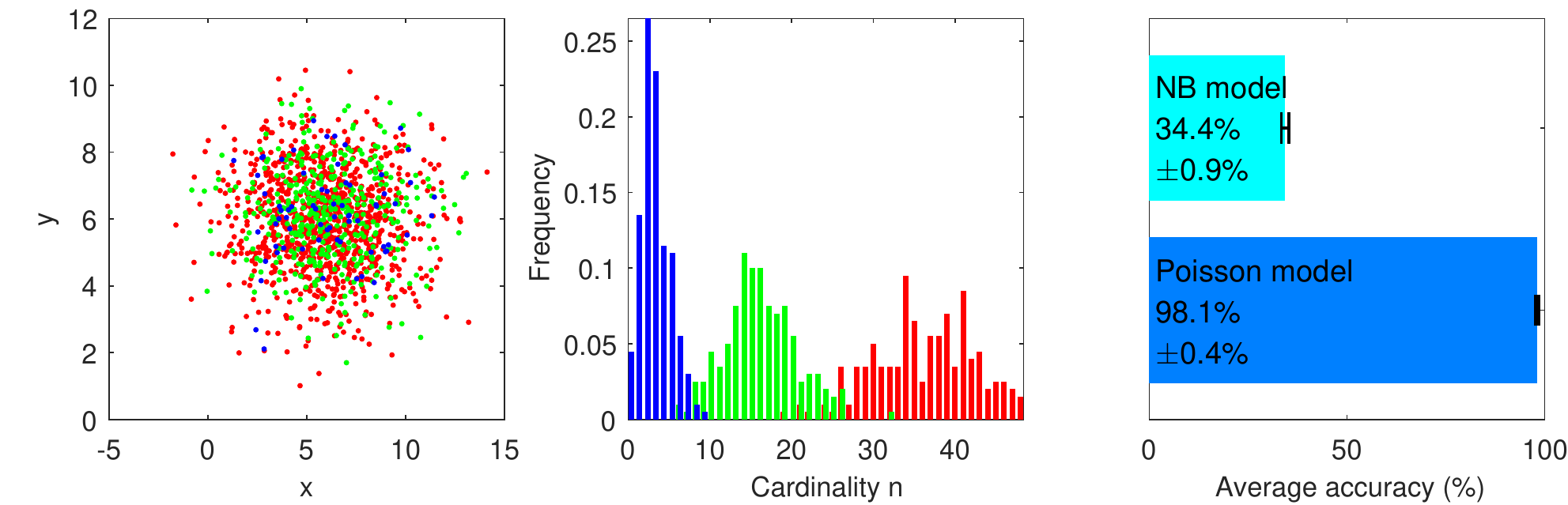}
\par\end{centering}
\begin{centering}
\vspace{1mm}
\par\end{centering}
\centering{}}\vspace{-2mm}
\par\end{centering}
\begin{centering}
\subfloat[\label{fig:Classifi_Sim_mix}Classes 2 and 3: overlap in feature,
well-separated in cardinality. Classes 1 and 2: overlap in cardinality,
well-separated in feature. ]{\begin{centering}
\includegraphics[bb=0bp 20mm 163bp 220bp,height=4.25cm]{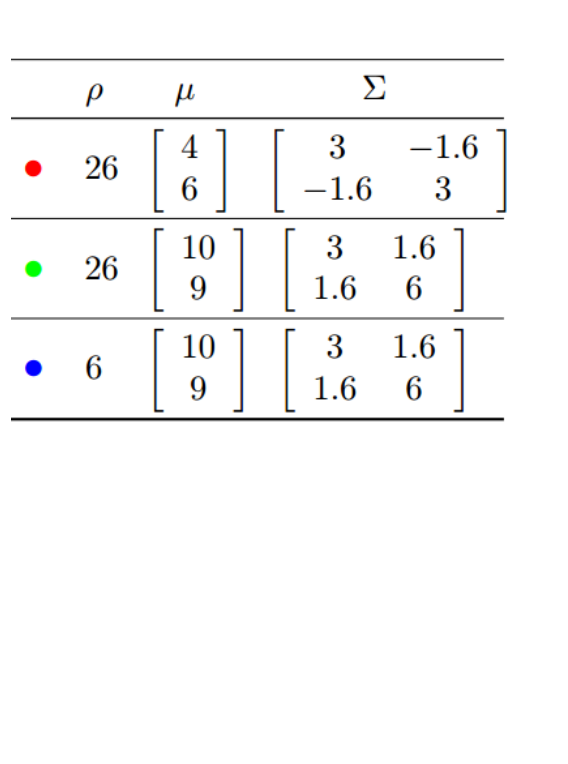}\includegraphics[height=4cm]{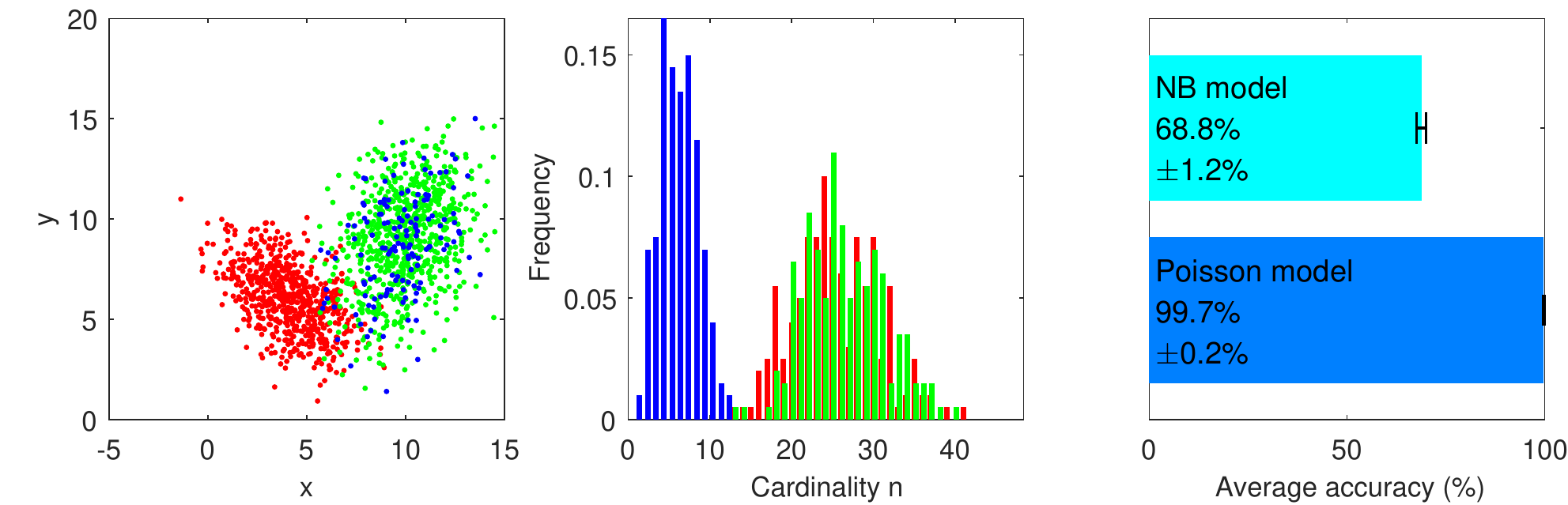}
\par\end{centering}
\centering{}\vspace{1mm}
}
\par\end{centering}
\caption{\label{fig:AP-sim}Model, data and classification accuracy (No. correct
classifications / No. of observations in the test set \cite{manning2008info_retrieval})
for three scenarios.}

\vspace{-0mm}
\end{figure*}

We consider three diverse scenarios, each comprising three classes
simulated from Poisson point processes with Gaussian intensities shown
in Fig. \ref{fig:AP-sim}. In scenario (a), point patterns from each
class are well-separated from other classes in feature, but significantly
overlapping in cardinality (see Fig. \ref{fig:Classifi_Sim_sepa_feat}).
In scenario (b), point patterns from each class are well-separated
from other classes in cardinality, but significantly overlapping in
feature (see Fig. \ref{fig:Classifi_Sim_overlap_feat}). Scenario
(c) is a mix of (a) and (b), where: point patterns from Class 1 are
well-separated from other classes in features, but significantly overlapping
with Class 2 in cardinality; and the point patterns from Classes 2
and 3 significantly overlap in feature, but well-separated in cardinality
(see Fig. \ref{fig:Classifi_Sim_mix}). 

The fully observed training dataset comprises 600 point patterns (200
per class) is used to learn the NB/Poisson model in which each class
is modeled by a Gaussian density/intensity. In the test phase, 10
different test sets each comprises 300 point patterns (100 per class)
are used. The average classification performance is reported in Fig.
\ref{fig:AP-sim}. Observe that in scenario (a), both the NB and Poisson
models perform equally well, while in scenarios (b) and (c) where
the point patterns overlap in feature, the Poisson model outperforms
the NB model since it exploits the cardinality information in the
point patterns. \vspace{-0mm}

\subsubsection{Classification on the Texture dataset\label{subsec:Classification-Texture}}

\begin{figure}[h]
\begin{centering}
\vspace{-0mm}
\par\end{centering}
\begin{centering}
\subfloat[\label{fig:Texture_examples}Example images (circles represent detected
SIFT keypoints).]{\begin{centering}
\includegraphics[width=0.32\columnwidth]{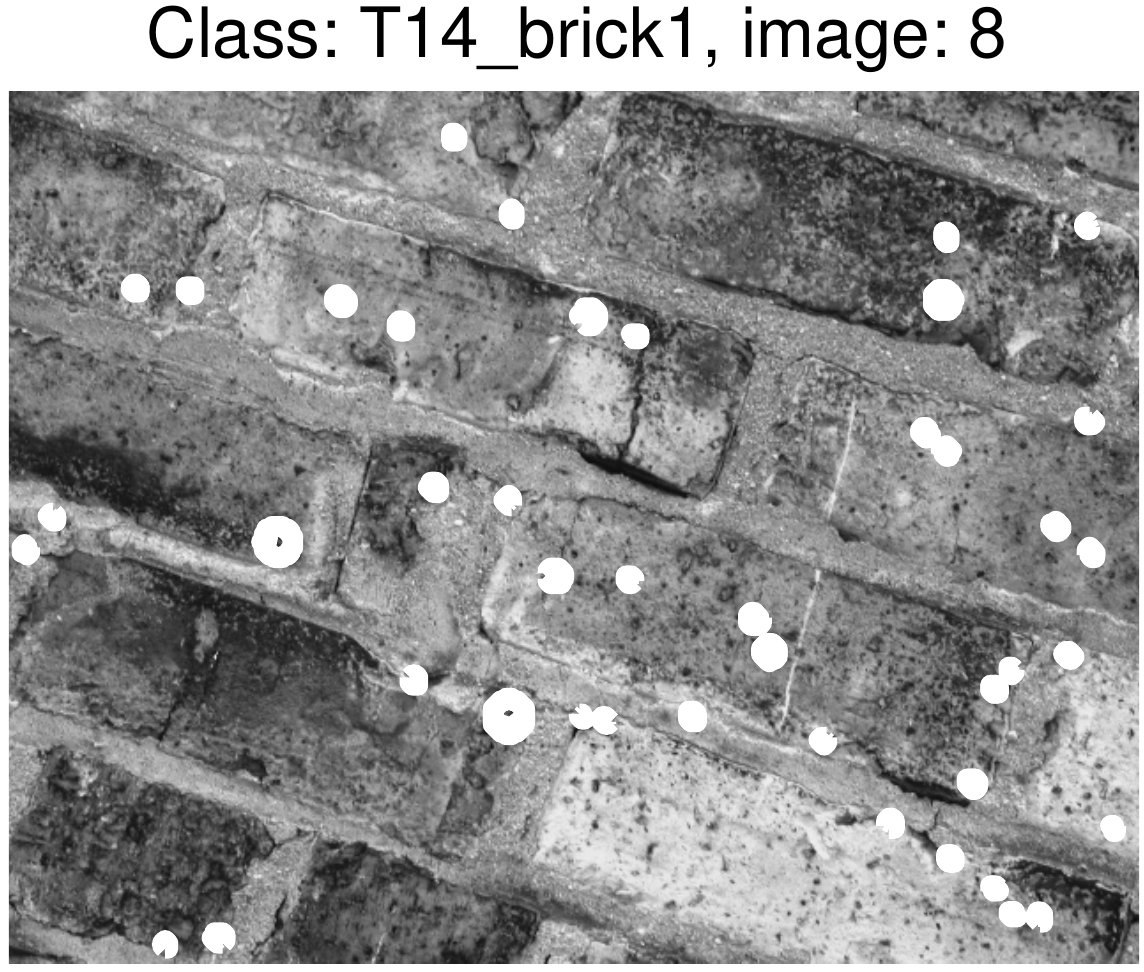}~\includegraphics[width=0.32\columnwidth]{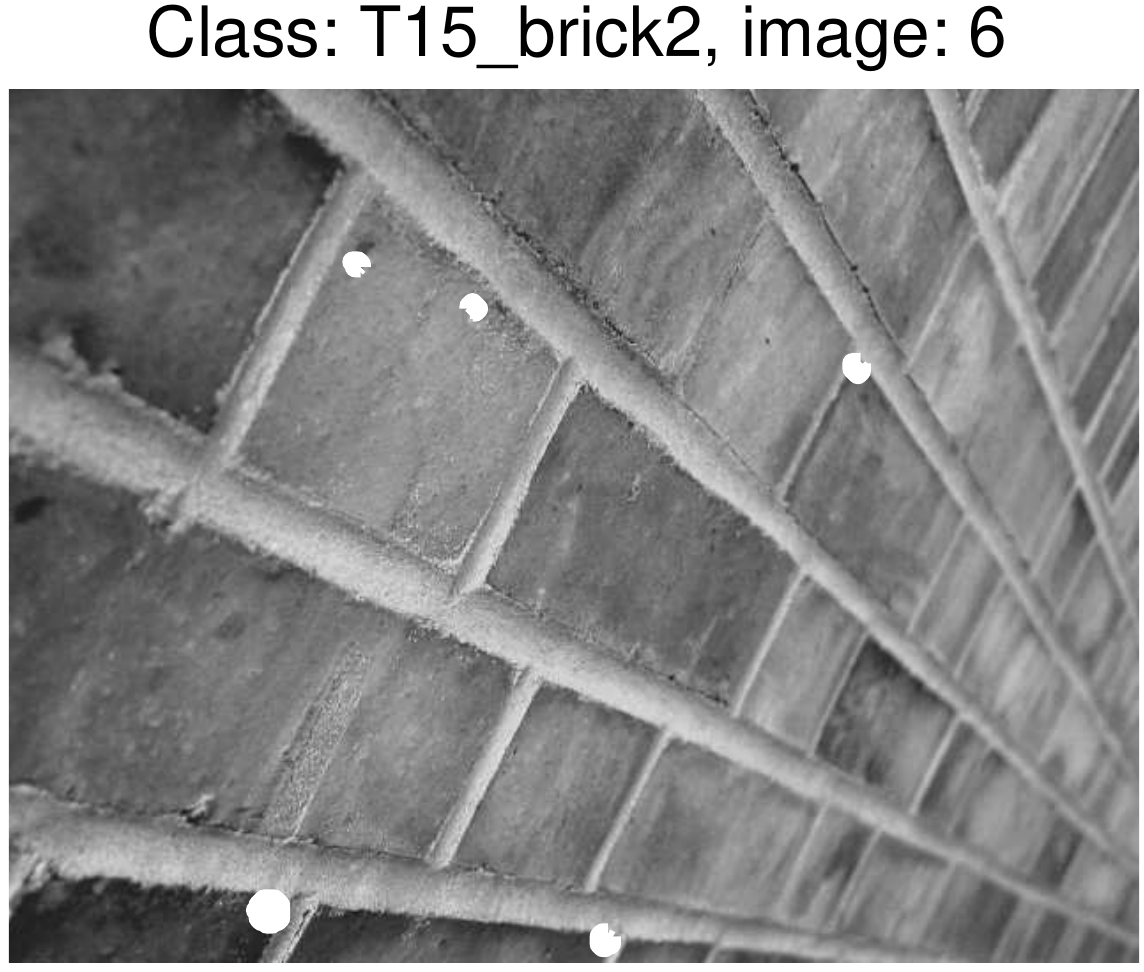}~\includegraphics[width=0.32\columnwidth]{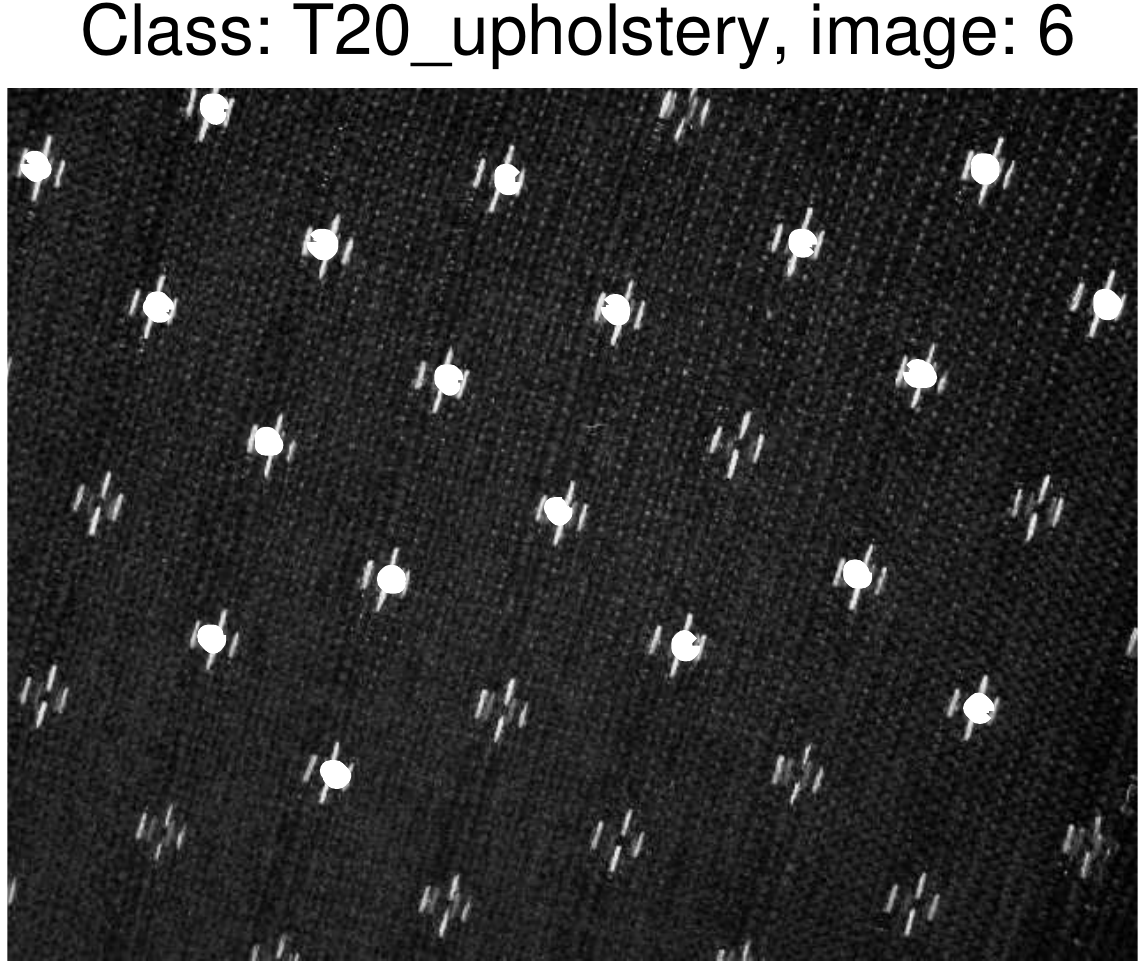}
\par\end{centering}
\begin{centering}
\vspace{1mm}
\par\end{centering}
}
\par\end{centering}
\begin{centering}
\subfloat[\label{fig:Texture_data}Extracted 2-D point patterns.]{\begin{centering}
\vspace{-2mm}
\par\end{centering}
\begin{centering}
\includegraphics[width=0.7\columnwidth]{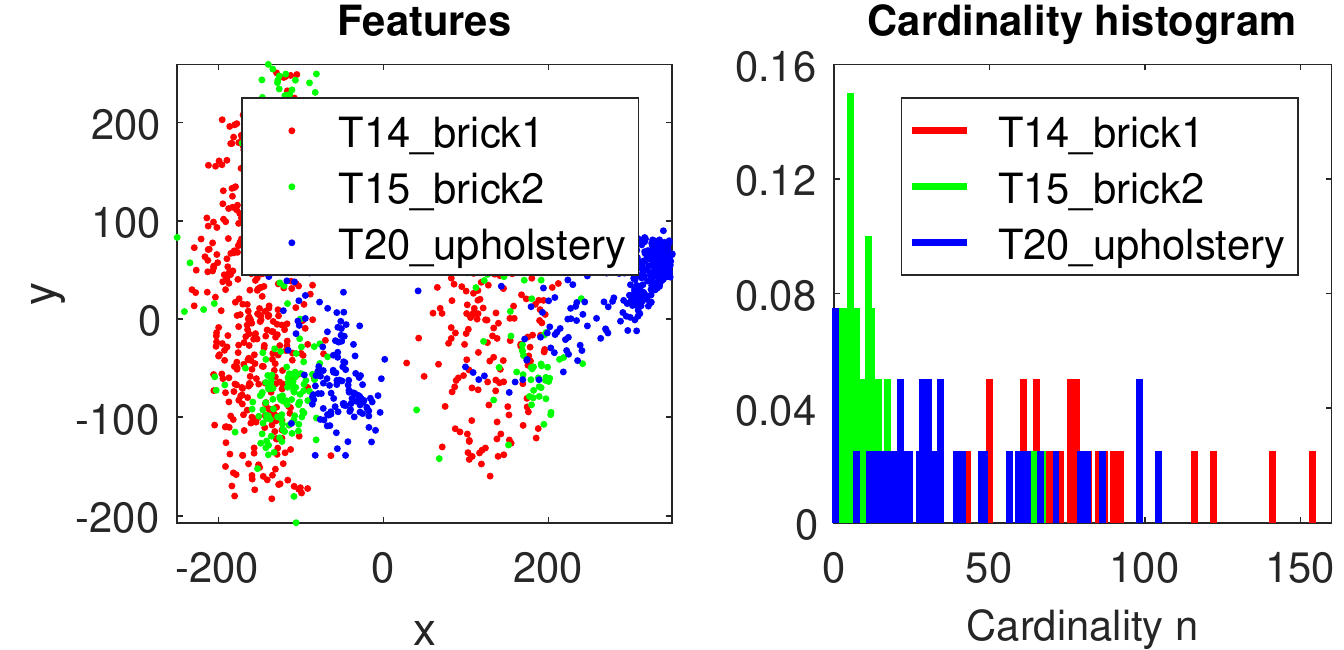}
\par\end{centering}
\vspace{1mm}

\vspace{-0mm}
}
\par\end{centering}
\caption{Three classes of the Texture dataset.}

\vspace{-0mm}
\end{figure}

Three classes ``T14 brick1'', ``T15 brick2'', and ``T20 upholstery''
of the Texture images dataset \cite{textureDataset} are considered.
Each class comprises 40 images, with some examples shown in Fig. \ref{fig:Texture_examples}.
Each image is processed by the SIFT algorithm (using the VLFeat library
\cite{vlfeatLib}) to produce a point pattern of 128-D SIFT features,
which is then compressed into a 2-D point pattern by Principal Component
Analysis (PCA). Fig. \ref{fig:Texture_data} shows the superposition
of the 2-D point patterns from the three classes along with their
cardinality histograms. 

\begin{figure}[h]
\begin{centering}
\vspace{-0mm}
\includegraphics[width=1\columnwidth]{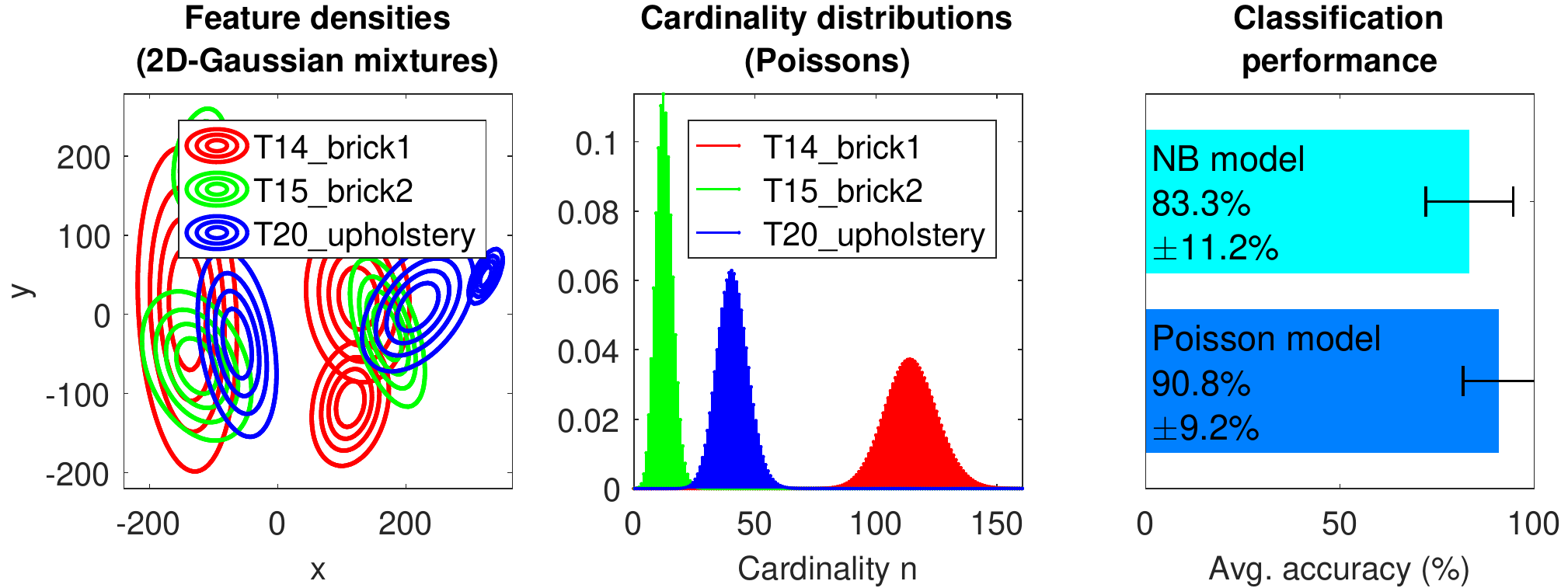}
\par\end{centering}
\vspace{-0mm}

\caption{\label{fig:Texture_classifi_dist_result} MLE of model parameters
and classification performance on the Texture dataset. The feature
densities are the same for both Poisson and NB models. The error-bars
represent standard deviations. }
\vspace{-0mm}
\end{figure}

A 4-fold cross validation scheme is used for performance evaluation.
In each fold, the fully observed training dataset comprising 30 images
per class is used to learn the NB/Poisson model in which each class
is parameterized by a 3-component Gaussian mixture density/intensity.
The test set comprises the remaining images (10 per class). Observe
from Fig \ref{fig:Texture_classifi_dist_result} that the Poisson
model outperforms NB, since it can exploit cardinality information
from the data. 

\subsection{Novelty Detection Experiments}

\begin{figure*}[t]
\vspace{-0mm}

\begin{centering}
\subfloat[\label{fig:novelty-sim-feature}Novelty well-separated from 'normal'
data in feature, but overlap in cardinality.]{\centering{}\includegraphics[bb=0bp 18mm 163bp 220bp,height=4.25cm]{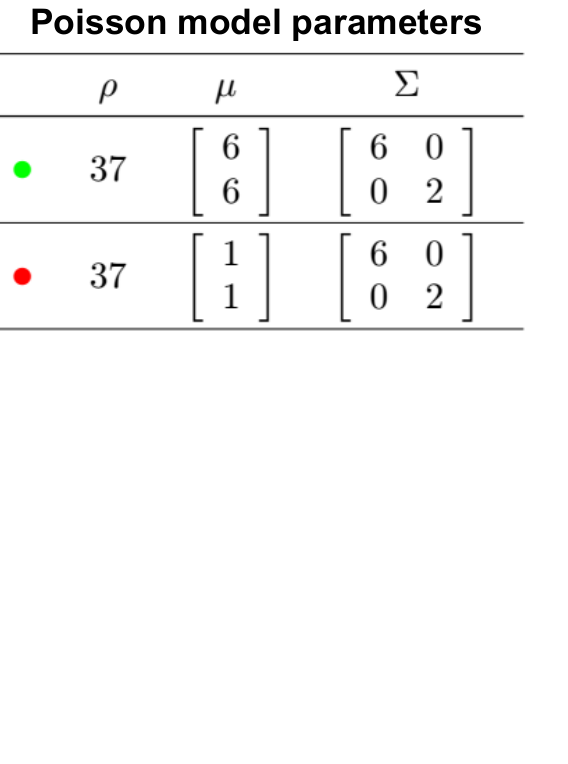}\includegraphics[height=4.2cm]{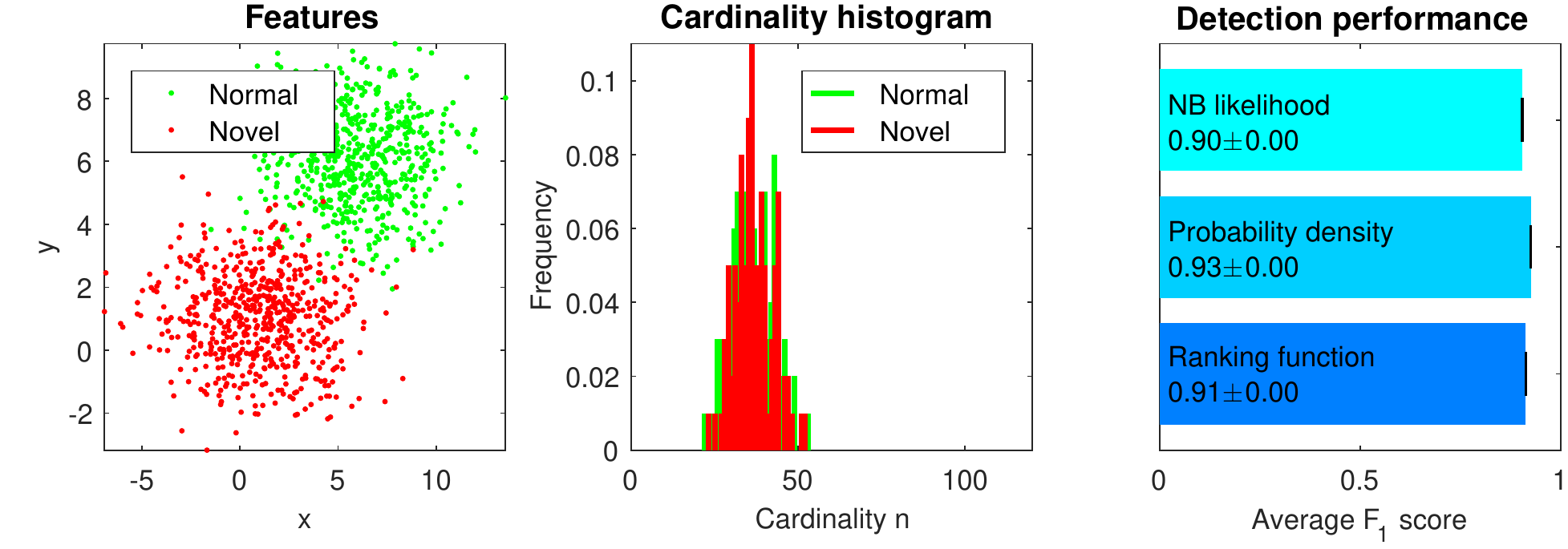}}
\par\end{centering}
\begin{centering}
\vspace{-2mm}
\subfloat[\label{fig:novelty-sim-mix}Novelty and 'normal' data partially overlap
in feature and cardinality.]{\centering{}\includegraphics[bb=0bp 20mm 163bp 220bp,height=4.25cm]{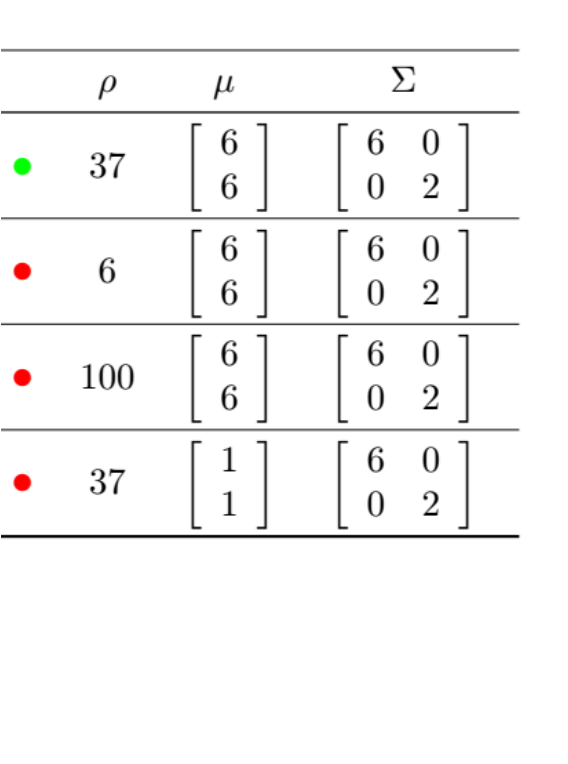}\includegraphics[height=4cm]{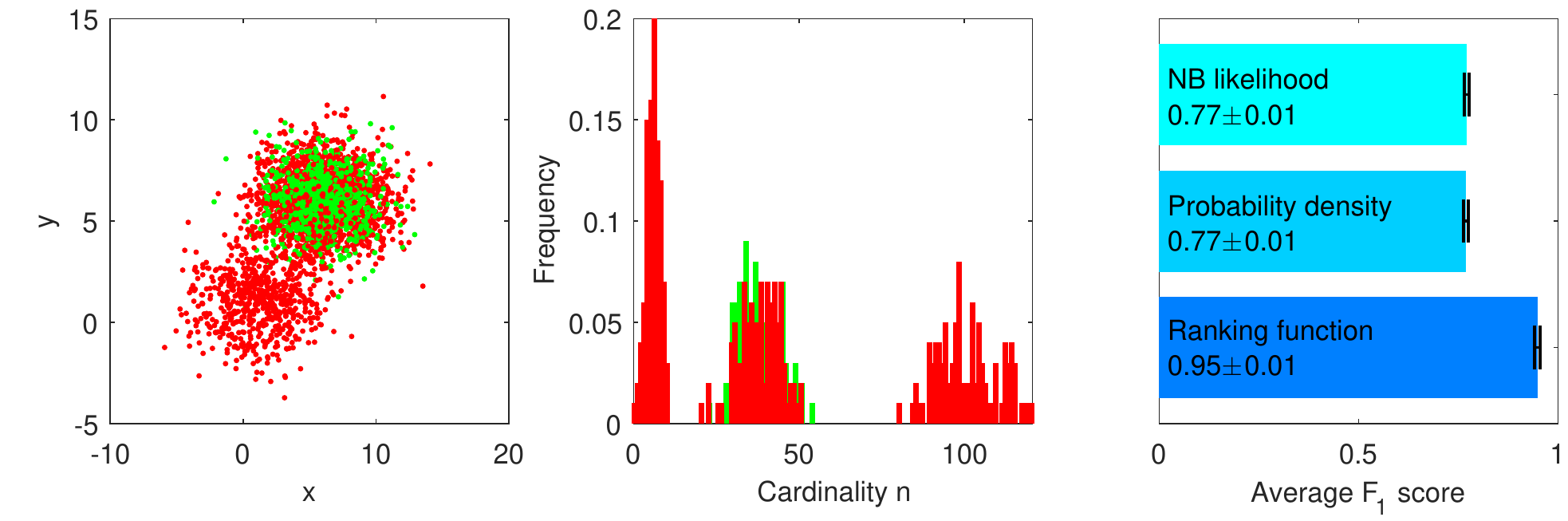}}
\par\end{centering}
\begin{centering}
\vspace{-2mm}
\subfloat[\label{fig:novelty-sim-low-card}Low-cardinality novelty that overlap
with 'normal' data in feature.]{\centering{}\includegraphics[bb=0bp 20mm 163bp 220bp,height=4.25cm]{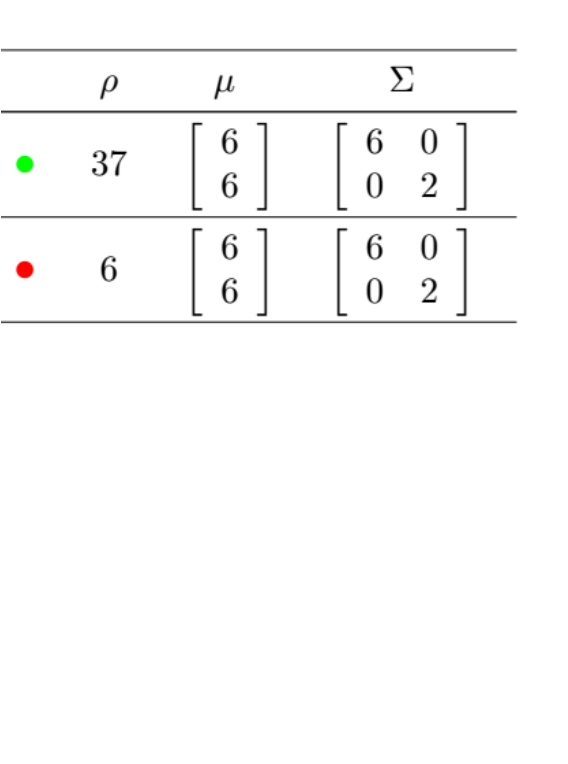}\includegraphics[height=4cm]{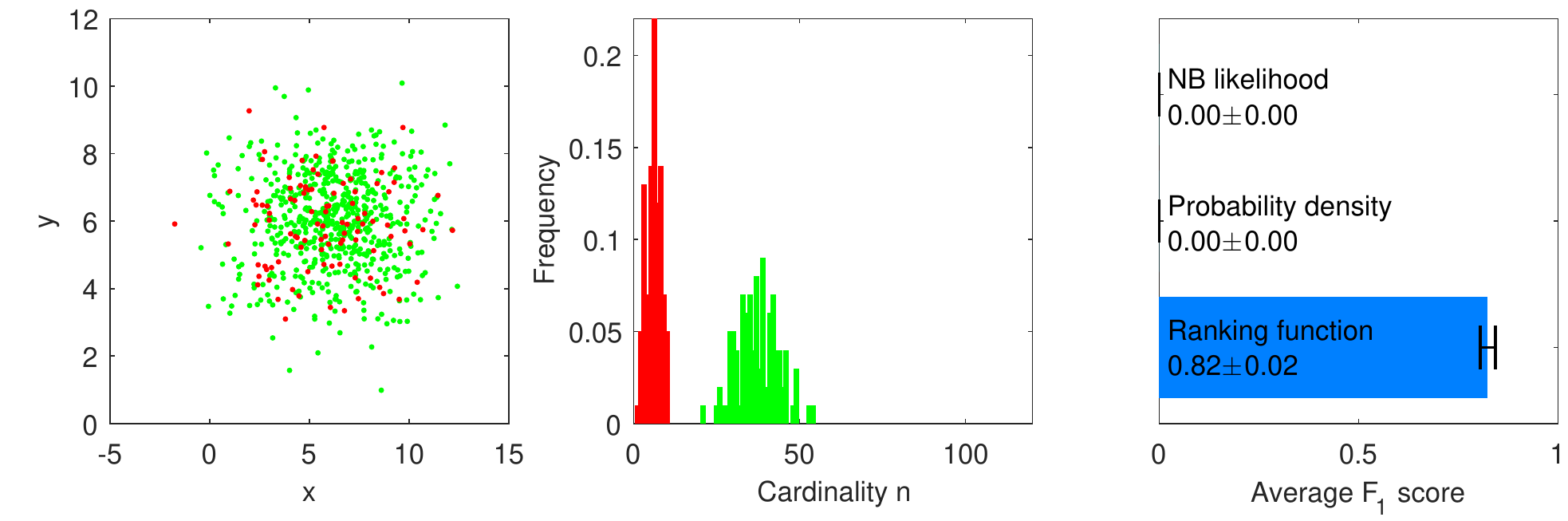}}
\par\end{centering}
\caption{\label{fig:novelty-sim}Model, data and novelty detection performance
for three scenarios. }

\vspace{-2mm}
\end{figure*}
This subsection presents two novelty detection experiments on both
simulated and real data using the Poisson model to illustrate the
effectiveness of the proposed ranking function against the NB likelihood
and standard probability density. Like the classification experiments,
ML is used to learn the parameters of the 'normal' NB and Poisson
models in the training phase. The novelty threshold is set at the
2nd 10-quantile of the ranking values of the 'normal' training data.
The detection performance measure is the $F_{1}$\emph{ score} \cite{manning2008info_retrieval}:
\[
F_{1}\mbox{(\mbox{precision},\mbox{ recall)}}\triangleq2\times\frac{\mbox{precision }\times\mbox{ recall}}{\mbox{precision }+\mbox{ recall}},
\]
where \textit{precision} is the proportion of correct decisions in
the output of the detector, and \textit{recall} is the proportion
of correctly identified novelties in the test set. To ensure functional
continuity of $F_{1}$, we define $F_{1}(0,0)\triangleq0$, i.e. its
limit at $(0,0)$.

\subsubsection{Novelty detection on simulated data}

We consider three simulated scenarios comprising 'normal' and novel
point patterns generated from Poisson point processes with 2-D Gaussian
intensities as shown in Fig. \ref{fig:novelty-sim}. All scenarios
have the same 'normal' point patterns, with cardinalities between
20 and 60. In scenario (a) novelties are well-separated from 'normal'
data in feature, but overlapping in cardinality (see Fig. \ref{fig:novelty-sim-feature}).
In scenario (b) novelties are overlapping with 'normal' data in feature,
but only partially overplapping in cardinality (see Fig. \ref{fig:novelty-sim-mix}).
In scenario (c) we remove the high cardinality novelties from (b)
(see Fig. \ref{fig:novelty-sim-low-card}).

In the training phase, the same 300 'normal' point patterns for each
scenario are used to learn the 'normal' NB/Poisson model that consists
of a Gaussian density/intensity. In the testing phase, 10 tests are
ran per scenario with each test set comprising 100 'normal' point
patterns and 100 novelties generated according to their respective
models. Observe from Fig. \ref{fig:novelty-sim-feature} that in scenario
(a) the NB likelihood, Poisson probability density, and Poisson ranking
function all perform well. Fig. \ref{fig:novelty-sim-mix} shows good
performance by the ranking function in scenario (b). The moderate
performance of the NB likelihood and probability density are inflated
by erroneously ranking high cardinality point patterns lower than
they are, due to the multiplication of many small numbers without
proper adjustment. Observe from Fig. \ref{fig:novelty-sim-low-card}
that after removing the high cardinality novelties in the test set,
only the ranking function perform well while the others fail. The
boxplots for test data in Fig. \ref{fig:Sim_anomaly_boxplot_low-card}
verified that only the proposed ranking function is consistent, whereas
the NB likelihood and the probability density even rank novelties
higher than 'normal' data.\vspace{-3mm}

\begin{flushleft}
\begin{figure}[h]
\vspace{0mm}

\begin{centering}
\subfloat[\label{fig:Sim_anomaly_boxplot_feature}]{\begin{centering}
\includegraphics[width=0.9\columnwidth]{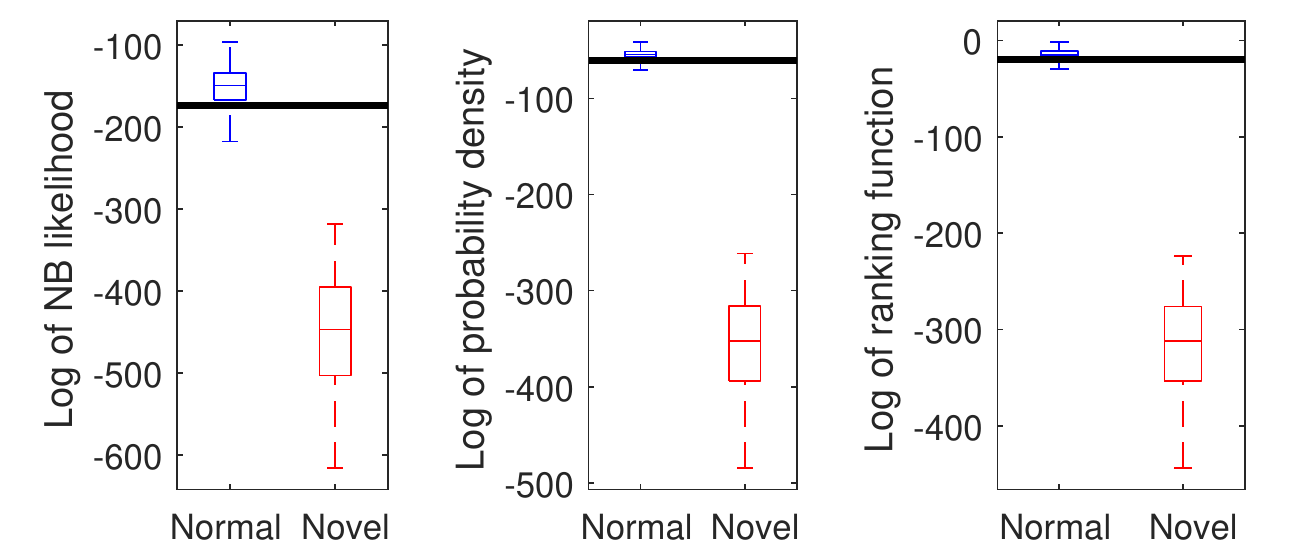}
\par\end{centering}
}
\par\end{centering}
\begin{centering}
\vspace{-1mm}
\subfloat[\label{fig:Sim_anomaly_boxplot_TOGETHER}]{\begin{centering}
\includegraphics[width=0.9\columnwidth]{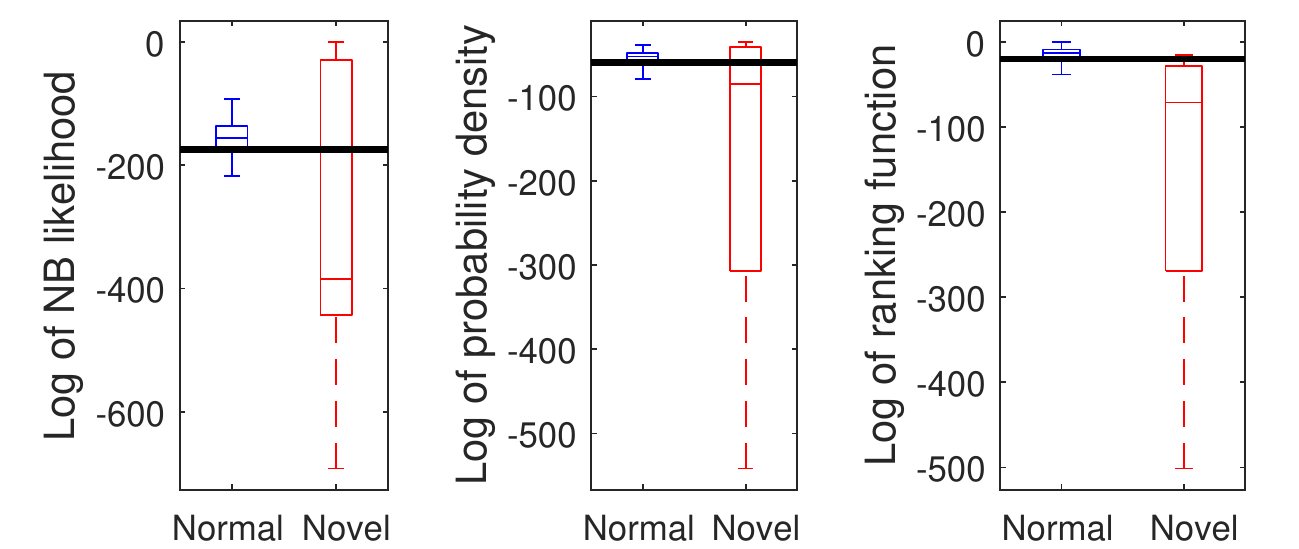}
\par\end{centering}
}
\par\end{centering}
\begin{centering}
\vspace{-1mm}
\subfloat[\label{fig:Sim_anomaly_boxplot_low-card}]{\begin{centering}
\includegraphics[width=0.9\columnwidth]{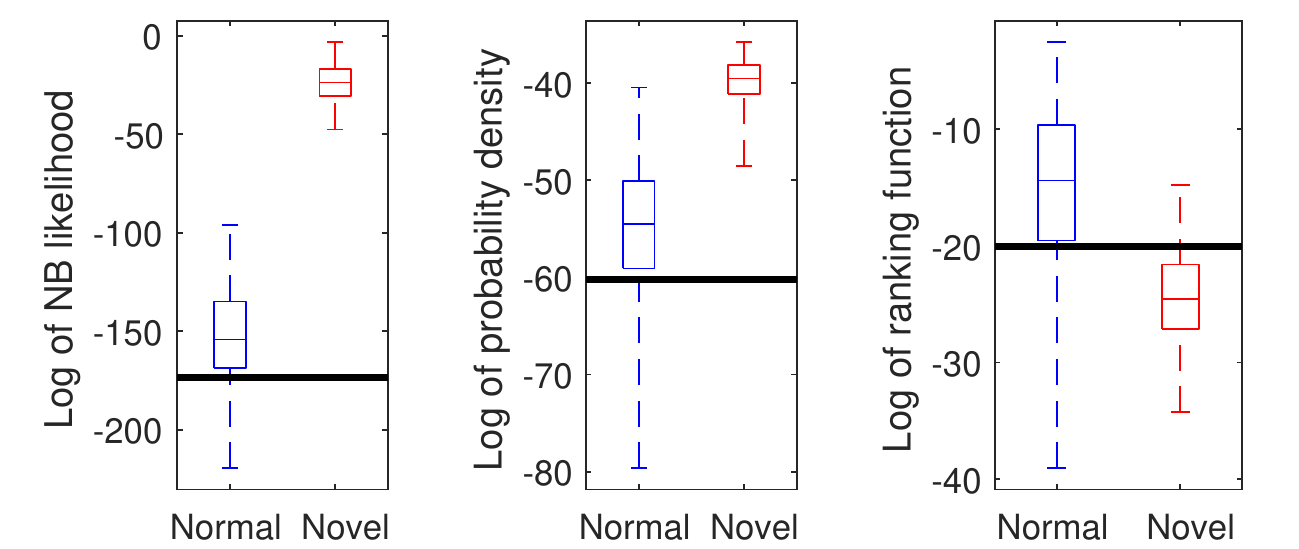}
\par\end{centering}
}
\par\end{centering}
\caption{Boxplots of: NB likelihood, probability density, and ranking function
for the three simulated scenarios in Fig. \ref{fig:novelty-sim} (solid
line through each graph indicates the novelty threshold).}
\vspace{0mm}
\end{figure}
\par\end{flushleft}

\subsubsection{Novelty detection on the Texture dataset}

For this experiment, data from class ``T14 brick1'' of the Texture
dataset from subsection \ref{subsec:Classification-Texture}, are
considered 'normal' while novel data are taken from class ``T20 upholstery''. 

\begin{figure}[h]
\begin{centering}
\includegraphics[width=0.38\columnwidth]{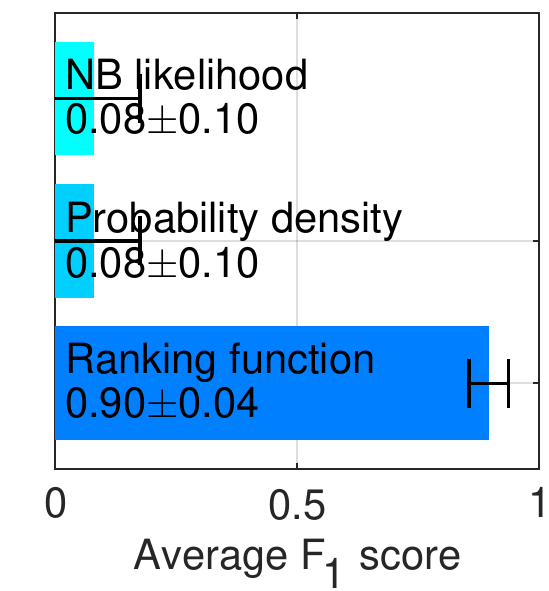}
\par\end{centering}
\vspace{-1mm}

\caption{\label{fig:Texture_anomaly_result}Averaged novelty detection performance
on the Texture dataset for: NB likelihood, probability density, and
proposed ranking function. The error-bars are standard deviations
of the F1-scores. }
\vspace{-1mm}
\end{figure}

\begin{figure}[h]
\begin{centering}
\includegraphics[width=0.9\columnwidth]{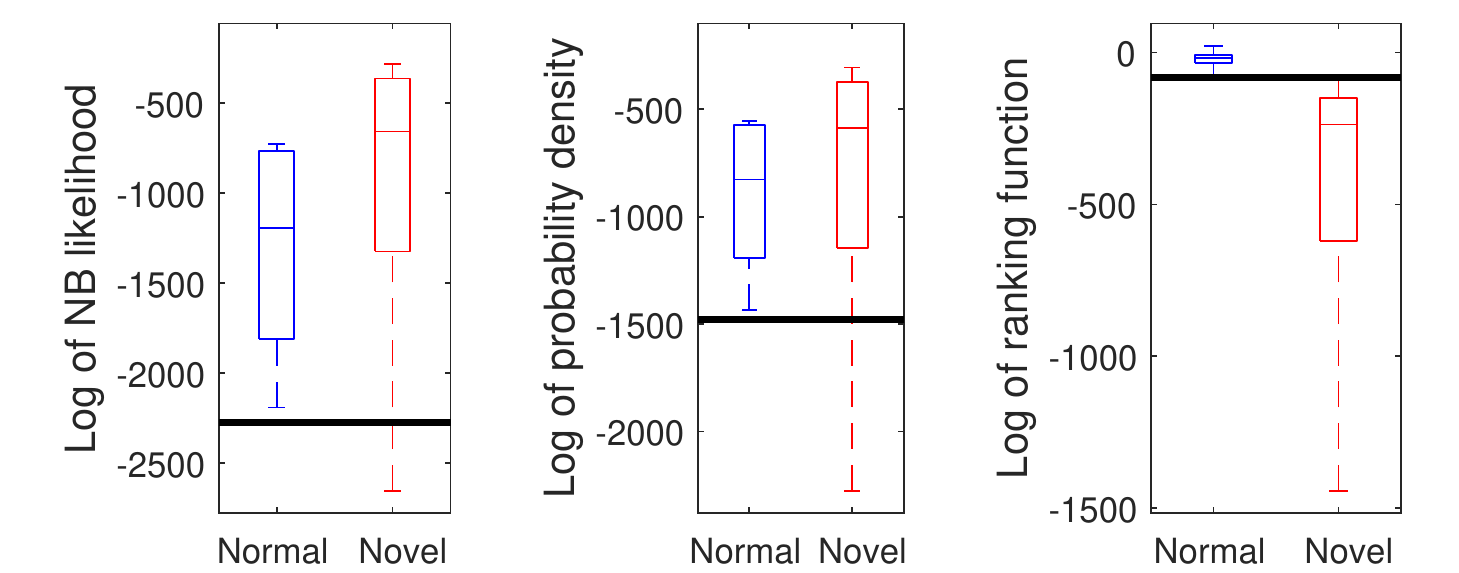}
\par\end{centering}
\caption{\label{fig:Texture_Boxplot}Boxplots of: NB likelihood; probability
density, and ranking function; for 'normal' and novel data in one
fold of the Texture dataset.}
\vspace{-1mm}
\end{figure}

A 4-fold cross validation scheme is used for performance evaluation.
In each fold, training data comprising 30 'normal' images is used
to learn the 'normal' NB/Poisson model that consists of a 3-component
Gaussian mixture density/intensity. The test set comprises the remaining
10 'normal' images and 10 novel images. The learned models are similar
to those of class ``T14 brick1'' in Fig. \ref{fig:Texture_classifi_dist_result}.
The novelty detection performance in Fig. \ref{fig:Texture_anomaly_result}
showed that ranking the data using the NB likelihood or the probability
density failed to detect most novelties, whereas the proposed ranking
function achieved a high F\textsubscript{1} score. Moreover, the
box plots for test data in Fig. \ref{fig:Texture_Boxplot} verified
that only the proposed ranking function provides a consistent ranking. 

\subsection{Clustering Experiments}

This subsection presents two clustering experiments with known number
of clusters using the EM clustering algorithm (outlined in subsection
\ref{subsec:EM-clustering}). For clustering performance measure,
we use Purity, normalized multual information (NMI), rand index and
$F_{1}$ score.

\subsubsection{EM clustering on simulated data \label{subsec:EM-sim}}

\begin{figure*}[t]
\begin{centering}
\vspace{-0mm}
\subfloat[All 3 clusters are well-separated from each other in feature, but
overlapping in cardinality.]{\centering{}~\includegraphics[height=4.75cm]{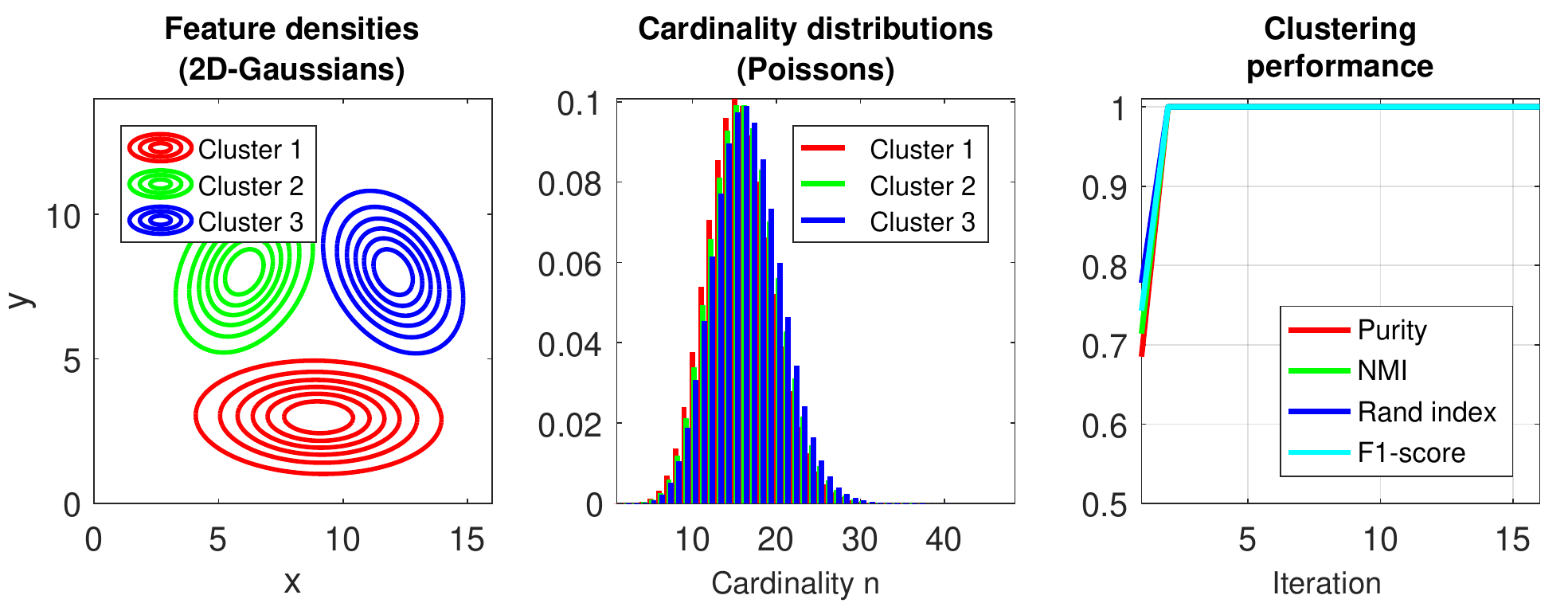}~~\includegraphics[bb=0bp 20mm 204bp 220bp,height=4.6cm]{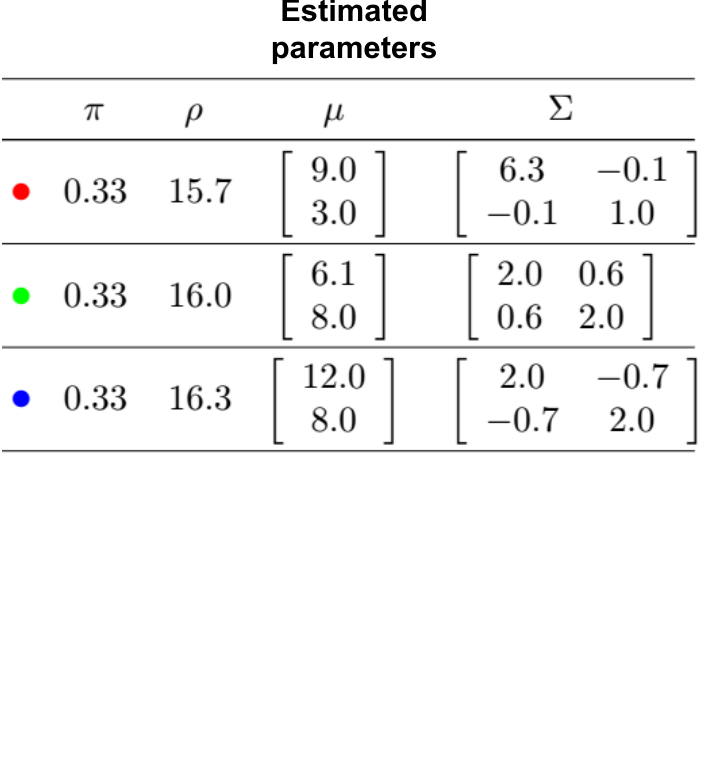}}
\par\end{centering}
\begin{centering}
\vspace{-2mm}
\subfloat[All 3 clusters are overlapping in feature, but well-separated from
each other in cardinality.]{\centering{}\includegraphics[height=4.27cm]{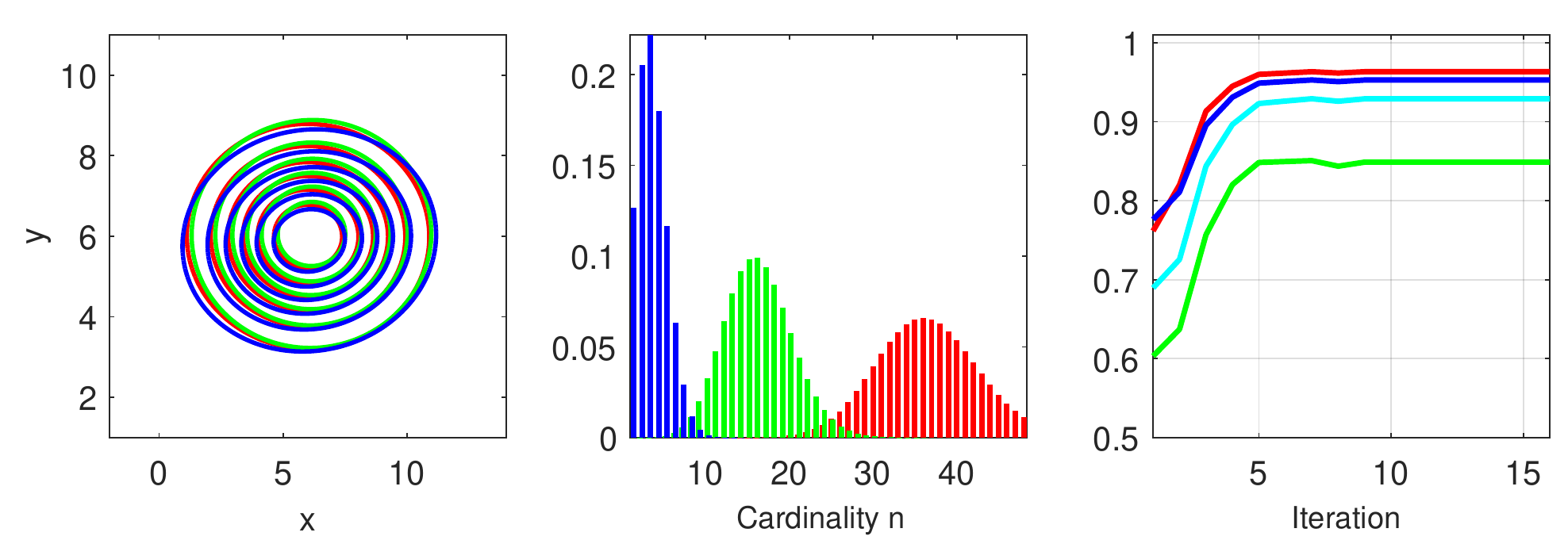}~~\includegraphics[bb=0bp 22mm 204bp 220bp,height=4.38cm]{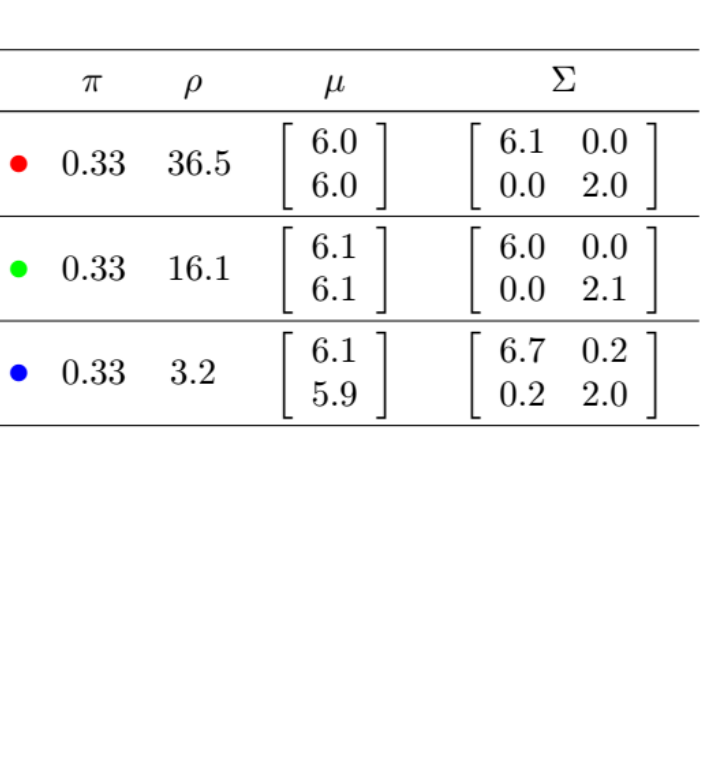}}
\par\end{centering}
\begin{centering}
\vspace{-2mm}
\subfloat[Clusters 2 and 3: overlap in feature, well-separated in cardinality.
Clusters 1 and 2: overlap in cardinality, well-separated in feature.]{\begin{centering}
\includegraphics[height=4.27cm]{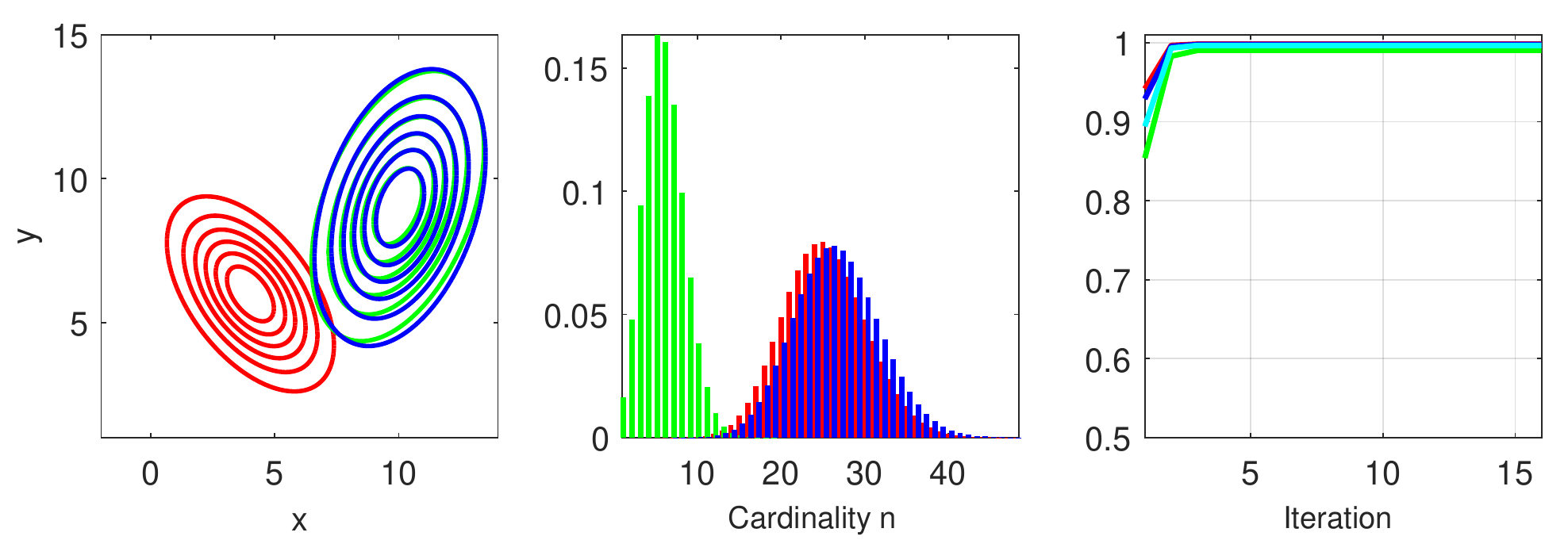}~~\includegraphics[bb=0bp 22mm 204bp 220bp,height=4.38cm]{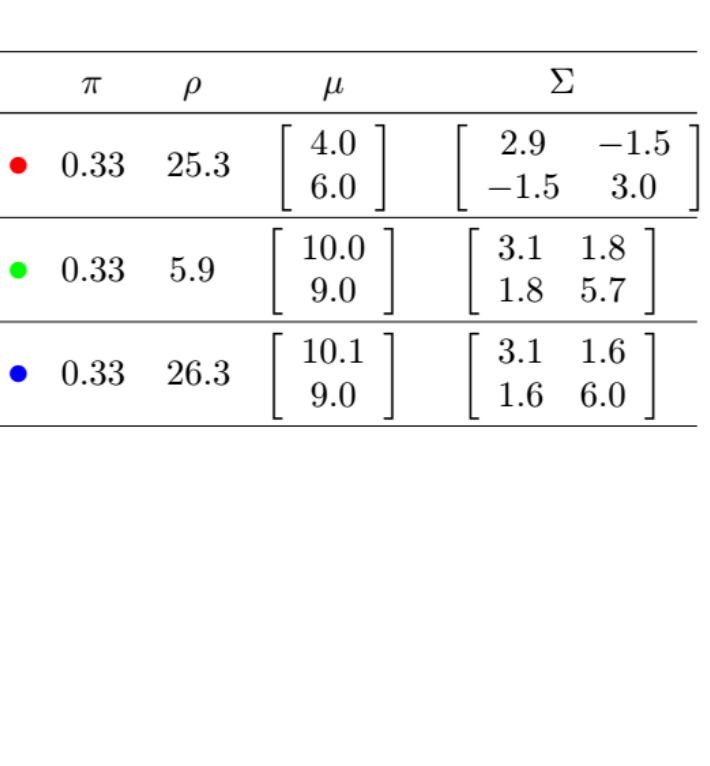}
\par\end{centering}
}
\par\end{centering}
\caption{\label{fig:EM_sim_dist_results}EM clustering performance on the three
simulated data scenarios in Fig. \ref{fig:AP-sim}.}

\vspace{-2mm}
\end{figure*}

This experiment uses the same simulated dataset described in section
\ref{subsec:sim-classifi} but without labels. Since there are three
clusters, we use a 3-component Poisson mixture model, where each constituent
Poisson point process is parameterized by a Gaussian intensity. The
clustering results in Fig. \ref{fig:EM_sim_dist_results} show that
the proposed point pattern clustering algorithm performs well on all
three scenarios. 

\subsubsection{EM clustering on the Texture dataset\label{subsec:EM-Texture}}

\begin{figure}[h]
\begin{centering}
\vspace{-0mm}
\includegraphics[width=1\columnwidth]{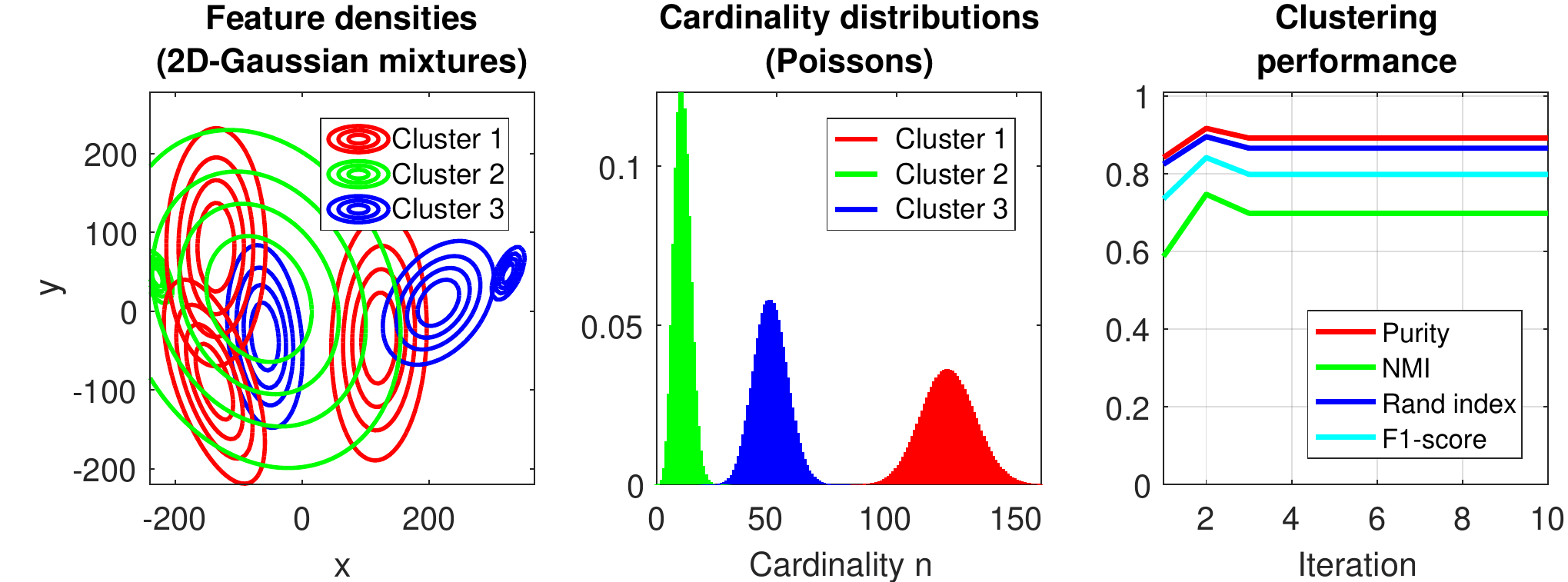}
\par\end{centering}
\caption{\label{fig:EM_Texture}EM clustering performance on the Texture dataset. }

\vspace{-2mm}
\end{figure}

This experiment uses the Texture dataset described in section \ref{subsec:Classification-Texture},
but without labels. Since there are three clusters, we use a 3-component
Poisson mixture model, where each constituent Poisson point process
is parameterized by a 3-component Gaussian mixture intensity (similar
to subsection \ref{subsec:Classification-Texture}). The M-step of
the proposed EM algorithm is accomplished by applying the standard
EM algorithm to find the data-weighted MLE of the Gausian mixture
parameter. The clustering results in Fig. \ref{fig:EM_Texture} show
that the proposed algorithm performs well on real data.

\section{Conclusions\label{sec:Conclusions}}

This article outlined a framework for model-based learning for point
pattern data using point process theory. In particular, we demonstrated
the use of point process models for various learning tasks. Our main
aim is to introduce an available toolset that facilitates research
in machine learning for point pattern data. While the utility of the
framework was only demonstrated on representative learning tasks such
as classification, novelty detection and clustering, such framework
is flexible enough to accommodate other learning tasks. For tractability,
the proposed algorithms are based on very simple models. Improved
performance on real data can be achieved with more sophisticated models,
albeit at higher computational costs. More complex datasets, where
the i.i.d. assumption is no longer adequate, require sophisticated
point process models such as Gibbs to capture interactions between
the elements of the point patterns. Developing efficient techniques
for learning such models is an active research area in statistics.

\bibliographystyle{IEEEtran}
\bibliography{Refs/ref1,Refs/ref_BaNgu,Refs/ref_BaTuong,Refs/ref_Dinh,Refs/Thuong}

\begin{thebibliography}{10}
\providecommand{\url}[1]{#1}
\csname url@samestyle\endcsname
\providecommand{\newblock}{\relax}
\providecommand{\bibinfo}[2]{#2}
\providecommand{\BIBentrySTDinterwordspacing}{\spaceskip=0pt\relax}
\providecommand{\BIBentryALTinterwordstretchfactor}{4}
\providecommand{\BIBentryALTinterwordspacing}{\spaceskip=\fontdimen2\font plus
\BIBentryALTinterwordstretchfactor\fontdimen3\font minus
  \fontdimen4\font\relax}
\providecommand{\BIBforeignlanguage}[2]{{%
\expandafter\ifx\csname l@#1\endcsname\relax
\typeout{** WARNING: IEEEtran.bst: No hyphenation pattern has been}%
\typeout{** loaded for the language `#1'. Using the pattern for}%
\typeout{** the default language instead.}%
\else
\language=\csname l@#1\endcsname
\fi
#2}}
\providecommand{\BIBdecl}{\relax}
\BIBdecl

\bibitem{amores2013multiple_intance_review}
J.~Amores, ``Multiple instance classification: Review, taxonomy and comparative
  study,'' \emph{Artificial Intelligence}, vol. 201, pp. 81--105, 2013.

\bibitem{foulds2010multi_instance_review}
J.~Foulds and E.~Frank, ``A review of multi-instance learning assumptions,''
  \emph{The Knowledge Engineering Review}, vol.~25, no.~01, pp. 1--25, 2010.

\bibitem{maron1961NB_setSizeVaried}
M.~E. Maron, ``Automatic indexing: an experimental inquiry,'' \emph{JACM},
  vol.~8, no.~3, pp. 404--417, 1961.

\bibitem{joachims1996probabilistic}
T.~Joachims, ``A probabilistic analysis of the rocchio algorithm with tfidf for
  text categorization.'' DTIC Document, Tech. Rep., 1996.

\bibitem{mccallum1998comparison_NBtextClassifi}
A.~McCallum and K.~Nigam, ``{A comparison of event models for naive Bayes text
  classification},'' in \emph{AAAI-98 Workshop learning for text
  categorization}, vol. 752, 1998, pp. 41--48.

\bibitem{csurka2004visual}
G.~Csurka, C.~Dance, L.~Fan, J.~Willamowski, and C.~Bray, ``Visual
  categorization with bags of keypoints,'' in \emph{Workshop statistical
  learning in computer vision, ECCV}, 2004.

\bibitem{fei2005bayesian}
L.~Fei-Fei and P.~Perona, ``{A Bayesian hierarchical model for learning natural
  scene categories},'' in \emph{IEEE Comput. Soc. Conf. Comput. Vision and
  Pattern Recognition (CVPR), 2005}, vol.~2.\hskip 1em plus 0.5em minus
  0.4em\relax IEEE, 2005, pp. 524--531.

\bibitem{chickering1999fast_sparse_data}
D.~M. Chickering and D.~Heckerman, ``Fast learning from sparse data,'' in
  \emph{Proc. 15th Conf. Uncertainty in artificial intelligence}.\hskip 1em
  plus 0.5em minus 0.4em\relax Morgan Kaufmann Publishers Inc., 1999, pp.
  109--115.

\bibitem{jing2007k_means_sparse_data}
L.~Jing, M.~K. Ng, and J.~Z. Huang, ``An entropy weighting k-means algorithm
  for subspace clustering of high-dimensional sparse data,'' \emph{IEEE Trans.
  Knowl. Data Eng.}, vol.~19, no.~8, pp. 1026--1041, 2007.

\bibitem{markou2003novelty_p1}
M.~Markou and S.~Singh, ``Novelty detection: a review -- part 1: statistical
  approaches,'' \emph{Signal Process.}, vol.~83, no.~12, pp. 2481--2497, 2003.

\bibitem{bishop2006pattern}
C.~M. Bishop, \emph{Pattern recognition and machine learning}.\hskip 1em plus
  0.5em minus 0.4em\relax Springer, 2006.

\bibitem{murphy2012machine}
K.~P. Murphy, \emph{Machine learning: a probabilistic perspective}.\hskip 1em
  plus 0.5em minus 0.4em\relax MIT press, 2012.

\bibitem{cadez2000EMclustering_VariableLengthData}
I.~V. Cadez, S.~Gaffney, and P.~Smyth, ``A general probabilistic framework for
  clustering individuals and objects,'' in \emph{Proc. 6th ACM SIGKDD Int.
  Conf. knowledge discovery and data mining}, 2000, pp. 140--149.

\bibitem{Stoyan95}
D.~Stoyan, W.~S. Kendall, and J.~Mecke, \emph{Stochastic geometry and its
  applications}.\hskip 1em plus 0.5em minus 0.4em\relax John Wiley \& Sons,
  1995.

\bibitem{Daley88}
D.~J. Daley and D.~Vere-Jones, \emph{An introduction to the theory of point
  processes}.\hskip 1em plus 0.5em minus 0.4em\relax Springer, 1988, vol.~2.

\bibitem{moller2003point_processes}
J.~Moller and R.~P. Waagepetersen, \emph{Statistical inference and simulation
  for spatial point processes}.\hskip 1em plus 0.5em minus 0.4em\relax CRC
  Press, 2003.

\bibitem{vo2016model-based_PP}
B.-N. Vo, Q.~N. Tran, D.~Phung, and B.-T. Vo, ``Model-based classification and
  novelty detection for point pattern data,'' in \emph{23rd Intl. Conf. Pattern
  Recognition (ICPR)}, Dec. 2016.

\bibitem{tran2016clustering_PP}
Q.~N. Tran, B.-N. Vo, D.~Phung, and B.-T. Vo, ``Clustering for point pattern
  data,'' in \emph{23rd Int. Conf. Pattern Recognition (ICPR)}, Dec 2016.

\bibitem{PV_RFSMODEL_FUSION14}
D.~Phung and B.-N. Vo, ``A random finite set model for data clustering,'' in
  \emph{Proc. 17th Annu. Conf. Inf. Fusion}, Salamanca, Spain, 2014.

\bibitem{illian2008statistical}
J.~Illian, A.~Penttinen, H.~Stoyan, and D.~Stoyan, \emph{Statistical analysis
  and modelling of spatial point patterns}.\hskip 1em plus 0.5em minus
  0.4em\relax John Wiley \& Sons, 2008.

\bibitem{baccelli_bk10}
F.~Baccelli and B.~Blaszczyszyn, \emph{Stochastic Geometry and Wireless
  Networks: Volume 1: Theory Foundation and Trends in Networking}.\hskip 1em
  plus 0.5em minus 0.4em\relax Now Publishers Inc, 2010, vol.~1.

\bibitem{mahler2014advances}
R.~P. Mahler, \emph{Advances in statistical multisource-multitarget information
  fusion}.\hskip 1em plus 0.5em minus 0.4em\relax Artech House, Inc., 2014.

\bibitem{lieshout2000markov}
M.~van Lieshout, \emph{Markov Point Processes and their Applications}.\hskip
  1em plus 0.5em minus 0.4em\relax Imperial College Press, 2000.

\bibitem{baddeley2007spatial}
A.~Baddeley, I.~B{\'a}r{\'a}ny, and R.~Schneider, ``Spatial point processes and
  their applications,'' \emph{Stochastic Geometry: Lectures given at the CIME
  Summer School held in Martina Franca, Italy, September 13--18, 2004}, pp.
  1--75, 2007.

\bibitem{geyer1999likelihood}
C.~J. Geyer \emph{et~al.}, ``Likelihood inference for spatial point
  processes,'' \emph{Stochastic geometry: likelihood and computation}, vol.~80,
  pp. 79--140, 1999.

\bibitem{vo2005sequential}
B.-N. Vo, S.~Singh, and A.~Doucet, ``Sequential monte carlo methods for
  multitarget filtering with random finite sets,'' \emph{Aerosp. Electron.
  Syst., IEEE Trans.}, vol.~41, no.~4, pp. 1224--1245, 2005.

\bibitem{Mahler_03}
R.~Mahler, ``Multi-target {Bayes} filtering via first-order multi-target
  moments,'' \emph{IEEE Trans. Aerospace \& Electronic Systems}, vol.~39,
  no.~4, pp. 1152--1178, 2003.

\bibitem{mahler2007statistical}
R.~P. Mahler, \emph{Statistical multisource-multitarget information
  fusion}.\hskip 1em plus 0.5em minus 0.4em\relax Artech House, Inc., 2007.

\bibitem{huttenlocher1993tracking_Hausdorff}
D.~P. Huttenlocher, J.~J. Noh, and W.~J. Rucklidge, ``Tracking non-rigid
  objects in complex scenes,'' in \emph{Proc. 4th Int. Conf. Comput. Vision,
  1993}.\hskip 1em plus 0.5em minus 0.4em\relax IEEE, 1993, pp. 93--101.

\bibitem{rockafellar2009variational}
R.~T. Rockafellar and R.~J.-B. Wets, \emph{Variational analysis}.\hskip 1em
  plus 0.5em minus 0.4em\relax Springer Sci. \& Business Media, 2009.

\bibitem{gavrila1999Chamfer_distance}
D.~M. Gavrila and V.~Philomin, ``Real-time object detection for "smart"
  vehicles,'' in \emph{Proc. 7th Int. Conf. Comput. Vision, 1999}, vol.~1,
  1999, pp. 87--93.

\bibitem{zhang2007EMD_kernel}
J.~Zhang, M.~Marsza{\l}ek, S.~Lazebnik, and C.~Schmid, ``Local features and
  kernels for classification of texture and object categories: A comprehensive
  study,'' \emph{Int. J. Comput. Vision}, vol.~73, no.~2, pp. 213--238, 2007.

\bibitem{rubner1998EarthMoversDistance}
Y.~Rubner, C.~Tomasi, and L.~J. Guibas, ``A metric for distributions with
  applications to image databases,'' in \emph{6th IEEE Int. Conf. Comput.
  Vision}, 1998, pp. 59--66.

\bibitem{ogata1984likelihood}
Y.~Ogata and M.~Tanemura, ``Likelihood analysis of spatial point patterns,''
  \emph{J. Royal Statistical Society. Series B (Methodological)}, pp. 496--518,
  1984.

\bibitem{geyer1994simulation}
C.~J. Geyer and J.~M{\o}ller, ``Simulation procedures and likelihood inference
  for spatial point processes,'' \emph{Scandinavian J. statistics}, pp.
  359--373, 1994.

\bibitem{takacs1986estimator}
R.~Takacs, ``Estimator for the pair--potential of a gibbsian point process,''
  \emph{Statistics: A J. Theoretical and Applied Statistics}, vol.~17, no.~3,
  pp. 429--433, 1986.

\bibitem{fiksel1988estimation}
T.~Fiksel, ``Estimation of interaction potentials of gibbsian point
  processes,'' \emph{Statistics}, vol.~19, no.~1, pp. 77--86, 1988.

\bibitem{besag1975statistical}
J.~Besag, ``Statistical analysis of non-lattice data,'' \emph{The
  statistician}, pp. 179--195, 1975.

\bibitem{besag1977some}
------, ``Some methods of statistical analysis for spatial data,''
  \emph{Bulletin of the Int. Statistical Institute}, vol.~47, no.~2, pp.
  77--92, 1977.

\bibitem{baddeley2000practical}
A.~Baddeley and R.~Turner, ``Practical maximum pseudolikelihood for spatial
  point patterns,'' \emph{Australian \& New Zealand J. Statistics}, vol.~42,
  no.~3, pp. 283--322, 2000.

\bibitem{jensen1991pseudolikelihood}
J.~L. Jensen and J.~M{\o}ller, ``Pseudolikelihood for exponential family models
  of spatial point processes,'' \emph{The Annals of Applied Probability}, pp.
  445--461, 1991.

\bibitem{pimentel2014review_novel_detect}
M.~A. Pimentel, D.~A. Clifton, L.~Clifton, and L.~Tarassenko, ``A review of
  novelty detection,'' \emph{Signal Process.}, vol.~99, pp. 215--249, 2014.

\bibitem{chandola2009anomaly}
V.~Chandola, A.~Banerjee, and V.~Kumar, ``Anomaly detection: A survey,''
  \emph{ACM Comput. Surveys (CSUR)}, vol.~41, no.~3, p.~15, 2009.

\bibitem{hodge2004survey}
V.~J. Hodge and J.~Austin, ``A survey of outlier detection methodologies,''
  \emph{Artificial Intelligence Review}, vol.~22, no.~2, pp. 85--126, 2004.

\bibitem{Jain1999data_clustering}
A.~K. Jain, M.~N. Murty, and P.~J. Flynn, ``Data clustering: a review,''
  \emph{ACM Comput. surveys (CSUR)}, vol.~31, no.~3, pp. 264--323, 1999.

\bibitem{russell2003artificial}
S.~Russell and P.~Norvig, \emph{Artificial Intelligence: A modern
  approach}.\hskip 1em plus 0.5em minus 0.4em\relax Prentice Hall, 2003.

\bibitem{jain2010clustering50yearsKmeans}
A.~K. Jain, ``Data clustering: 50 years beyond k-means,'' \emph{Pattern
  recognition letters}, vol.~31, no.~8, pp. 651--666, 2010.

\bibitem{xu2005survey_clustering}
R.~Xu and D.~Wunsch, ``Survey of clustering algorithms,'' \emph{IEEE Trans.
  Neural Networks}, vol.~16, no.~3, pp. 645--678, 2005.

\bibitem{zhang2009MIClustering}
M.-L. Zhang and Z.-H. Zhou, ``Multi-instance clustering with applications to
  multi-instance prediction,'' \emph{Appl. Intell.}, vol.~31, no.~1, pp.
  47--68, 2009.

\bibitem{zhang2009m3icClustering}
D.~Zhang, F.~Wang, L.~Si, and T.~Li, ``M3ic: Maximum margin multiple instance
  clustering,'' in \emph{IJCAI}, vol.~9, 2009, pp. 1339--1344.

\bibitem{dempster1977em_algorithm}
A.~P. Dempster, N.~M. Laird, and D.~B. Rubin, ``Maximum likelihood from
  incomplete data via the em algorithm,'' \emph{J. Royal Statistical Soc.
  Series B (Methodological)}, pp. 1--38, 1977.

\bibitem{Little2002StatisMissingData_EM}
R.~J. Little and D.~B. Rubin, \emph{Statistical analysis with missing
  data}.\hskip 1em plus 0.5em minus 0.4em\relax John Wiley \& Sons, 2002.

\bibitem{bernardo2009bayesian}
J.~M. Bernardo and A.~F. Smith, \emph{{B}ayesian theory}.\hskip 1em plus 0.5em
  minus 0.4em\relax John Wiley \& Sons, 2009, vol. 405.

\bibitem{Gelman_etal_bk03_Bayesian}
A.~Gelman, J.~Carlin, H.~Stern, and D.~Rubin, \emph{Bayesian Data
  Analysis}.\hskip 1em plus 0.5em minus 0.4em\relax Chapman \& Hall/CRC, 2003.

\bibitem{bilmes1998gentle_EM}
J.~A. Bilmes, ``A gentle tutorial of the em algorithm and its application to
  parameter estimation for gaussian mixture and hidden markov models,''
  \emph{Int. Comput. Sci. Institute}, vol.~4, no. 510, 1998.

\bibitem{Ghosh_bk03_Bayesian}
J.~Ghosh and R.~Ramamoorthi, \emph{Bayesian Nonparametrics}.\hskip 1em plus
  0.5em minus 0.4em\relax Springer Verlag, 2003.

\bibitem{Hjort_etal_bk10_bayesian}
N.~Hjort, C.~Holmes, P.~M{\"u}ller, and S.~Walker, \emph{Bayesian
  nonparametrics}.\hskip 1em plus 0.5em minus 0.4em\relax Cambridge Univ.
  Press, 2010.

\bibitem{Lin_etal_10construction}
D.~Lin, E.~Grimson, and J.~Fisher, ``Construction of dependent dirichlet
  processes based on poisson processes,'' \emph{Advances in Neural Information
  Processing Systems}, 2010.

\bibitem{Jordan_10hierarchical}
M.~Jordan, ``Hierarchical models, nested models and completely random
  measures,'' in \emph{Frontiers of Statistical Decision Making and Bayesian
  Analysis: In Honor of James O. Berger}.\hskip 1em plus 0.5em minus
  0.4em\relax Springer-Verlag, New York, NY, 2010.

\bibitem{Blackwell_MacQueen_73ferguson}
D.~Blackwell and J.~MacQueen, ``{F}erguson distributions via {P}{\'o}lya urn
  schemes,'' \emph{The annals of statistics}, vol.~1, no.~2, pp. 353--355,
  1973.

\bibitem{textureDataset}
S.~Lazebnik, C.~Schmid, and J.~Ponce, ``A sparse texture representation using
  local affine regions,'' \emph{IEEE Trans. Pattern Anal. Mach. Intell.},
  vol.~27, no.~8, pp. 1265--1278, 2005.

\bibitem{manning2008info_retrieval}
C.~D. Manning, P.~Raghavan, and H.~Sch{\"u}tze, \emph{Introduction to
  information retrieval}.\hskip 1em plus 0.5em minus 0.4em\relax Cambridge
  univ. press Cambridge, 2008, vol.~1.

\bibitem{vlfeatLib}
A.~Vedaldi and B.~Fulkerson, ``Vlfeat: An open and portable library of comput.
  vision algorithms,'' http://www.vlfeat.org/, 2008.

\end{thebibliography}

\end{document}